\documentclass[]{fairmeta}
\usepackage{amsmath}
\usepackage{amssymb}   % AMS symbol fonts, as in the AAAI supplement

\usepackage{tikz}
\usetikzlibrary{positioning,arrows.meta,calc,fit,backgrounds}

\usepackage{algorithm}
\usepackage{algpseudocode}
\usepackage{listings}
\usepackage{array}   % >{\raggedright\arraybackslash}p{} in the wide text tables
\usepackage{xspace}
\usepackage{needspace}
\newcommand{\slugurl}[1]{\begingroup\def\UrlBreaks{\do\.\do\/\do\-}\url{#1}\endgroup}

\DeclareFontShape{T1}{optimistic}{m}{sc}{<-> s * [0.88] assets/optimistic}{}
\DeclareFontShape{T1}{optimistic}{b}{sc}{<-> s * [0.88] assets/optimistic}{}
\DeclareFontShape{T1}{optimistic}{m}{it}{<-> s * [0.88] assets/optimistic}{}
\DeclareFontShape{T1}{optimistic}{b}{it}{<-> s * [0.88] assets/optimistic}{}
\DeclareFontShape{T1}{optimistic}{m}{sl}{<-> s * [0.88] assets/optimistic}{}
\DeclareFontShape{T1}{optimistic}{b}{sl}{<-> s * [0.88] assets/optimistic}{}
\DeclareFontShape{T1}{cmr}{m}{scit}{<->ssub * cmr/m/scsl}{}

\newcommand{\ndcg}{NDCG@10\xspace}
\DeclareRobustCommand{\re}{\textsc{RankEvolve}\xspace}
\DeclareRobustCommand{\execbench}{\textsc{ExecML}\xspace}

\title{RankEvolve: A Reliable Multi-Agent Auto-Research Harness for Evolving
Ranking Models}

\author{Zheng Chen}
\author{Linfeng Liu}
\author{Hong Li}
\author{Hong Yan}

\affiliation{Meta}

\abstract{Auto-research agents---LLM systems that evolve a machine-learning (ML) model
by proposing, implementing, training, and evaluating changes across
iterations---promise to automate applied ML's experimental loop.  Over such
long horizons, \emph{execution accuracy} is a binding constraint in our
setting: a change
can silently leak held-out data, omit a layer norm, disconnect a gradient, or
leave a train/eval flag unwired---one such defect burns hours of accelerator
time and yields an invalid metric---and over many iterations the damage
compounds while the run drifts off its prescribed process.  We
present \re{}, an auto-research framework that evolves a generative
ranking model end to end, with correctness engineered through two
disciplines.
An \emph{Executable Operating Protocol} (EOP) declares the workflow's phases,
gates, branches, and loops once, and the runtime enforces the compiled state
machine: the model performs each step, but the framework controls the process.
A \emph{meta-meta-harness} treats the agentic system as an execution graph
whose nodes are not raw model calls but complete, black-box coding-agent
products (e.g., Claude Code, Codex)---each itself a harness over a
model---that cross-check one another's work; in a budget-matched evaluation
this cross-checking lifts all-oracle execution accuracy from the best
single-product $45.8\%$ to $62.5\%$---the lever that makes long-horizon
iteration reliable.  Beneath both disciplines, an implemented
\emph{knowledge layer} carries findings---including negative
results---across iterations.  Deployed for twelve iterations on the
open-source HSTU recommender, \re{} reported \ndcg{} 0.2192 on MovieLens-20M
LARGE ($+4.48\%$ over the published anchor) and 0.1948 on BASE ($+2.80\%$),
with first-class negative results and leakage incidents.  Those incidents
seed \execbench{}-HSTU, the oracle benchmark behind that contrast: the Claude Code$+$Codex
pair beats every budget-matched single-product baseline (paired $+16.7$,
95\% CI $[6.6,26.7]$) at a $10.4\%$ silent critical-defect rate.  An equally
controlled LitGPT split, reported whatever its outcome, replicates the effect
beyond recommenders ($+12.5$, 95\% CI $[3.0,22.0]$); a paired ablation holds
the runtime fixed and toggles only per-step versus full-protocol injection.
The result is a falsifiable account of when runtime-controlled composition of
coding-agent products buys execution accuracy---and when it does not.
}

\date{August 1, 2026}

\hypersetup{
  pdftitle={RankEvolve: A Reliable Multi-Agent Auto-Research Harness for
    Evolving Ranking Models},
  pdfauthor={Zheng Chen, Linfeng Liu, Hong Li, Hong Yan},
}

\begin{document}

\maketitle

\section{Introduction}
\label{sec:introduction}

Applied ML research is a stylized loop: read the code and the literature,
form a hypothesis, implement it, launch training, validate the evaluator,
interpret the result, and decide what to try next.  The bottleneck is rarely
raw compute alone; it is the researcher who must sit in the loop at every
step.  LLM-driven systems for program search and scientific
discovery---FunSearch \citep{romera-paredes2024funsearch}, AlphaEvolve
\citep{novikov2025alphaevolve}, the AI Scientist line
\citep{lu2024aiscientist,yamada2025aiscientistv2}---show that much of the
loop is automatable, with the most reliable early successes in settings with
\emph{cheap, deterministic} evaluators.  A newer wave of ML-engineering
agents pushes the same loop toward real training pipelines
\citep{jiang2025aide,nam2025mlestar,toledo2025aira}.  Industrial teams now run
it on their own proprietary recommendation and ranking models
\citep{wang2026selfevolvingrec,kumar2026rea}, and academic work evolves compact
open recommenders such as NCF and SASRec \citep{kim2026selfevolverec}.  We know
of no established account, however, of long-horizon, code-level evolution of an
open generative recommender at the scale of HSTU \citep{zhai2024hstu}---deep
stack, GPU-hour training---checked against its published reference numbers.

Deploying the loop at that scale surfaces three challenges.
\textbf{(i)~Costly evolution.}  Each candidate change needs hours of
distributed GPU training, so only a few candidates can be evaluated per
iteration and every invalid run is expensive.  \textbf{(ii)~Fragile
execution.}  Even frontier coding agents inject subtle defects into mature ML
code---test-set leakage, a missing layer norm, a disconnected gradient, an
unwired feature flag, a platform-specific config error\footnote{For example,
NeMo/Megatron shifts next-token labels during \emph{data} processing
(\texttt{tokens}\,$=$\,\texttt{text[:-1]},
\texttt{labels}\,$=$\,\texttt{text[1:]}), not in the loss as Hugging Face
does; unshifted labels silently degenerate training into
\emph{self}-prediction with a deceptively low loss.}---that pass shallow CI yet invalidate the
metric, so the loop burns accelerator time and, worse, learns from illusory
outcomes \citep{chen2025mlrbench,luo2025aipitfalls}.
\textbf{(iii)~Long-horizon drift.}  In a pilot audit of a full-context
playbook, adherence to the prescribed process decayed sharply as context
accumulated (Figure~\ref{fig:faithfulness})---the drift a runtime-enforced
protocol removes.

The control abstractions these challenges call for are not missing: StateFlow formulates LLM
workflows as state machines \citep{wu2024stateflow}; GPTSwarm treats agents as
optimizable computational graphs \citep{zhuge2024gptswarm}; ADAS and AFlow
search over code-represented agents and workflows
\citep{hu2025adas,zhang2025aflow}; and LangGraph supplies a durable stateful
graph runtime \citep{langchain2026langgraph}.  Together these make ``agents are
graphs'' or ``workflows are state machines'' untenable as novel
contributions (Section~\ref{sec:related_work}), and we claim novelty for none
of them---nor for checkpointing, human-in-the-loop control, multi-agent
debate, or automatic agent design.  The gap is at the boundary between
authoring and execution: research procedures are authored as versioned
natural-language playbooks, whereas reliable long-horizon execution needs
explicit state, recovery, gates, and traceability; and a work node may need to invoke an entire coding-agent
\emph{product} rather than a single model call.  If two such products fail
differently, a review--repair edge can catch what either alone misses; if
their errors are correlated, composition only multiplies cost.  Existing work
does not establish that product-level composition buys objectively measured
correctness on a mature ML codebase at matched budget.  The recommender loops
above do not settle it either: they report model gains, on proprietary models
\citep{wang2026selfevolvingrec,kumar2026rea} or compact open ones
\citep{kim2026selfevolverec}, and none, to our knowledge, measures patch
correctness against an oracle or composes different coding-agent products.
The primary scientific question is therefore: \emph{does this composition boundary buy
execution accuracy, beyond spending the same budget on a single product?}

\paragraph{Contributions.}\mbox{}\\
\re{},\footnote{Distinct from the like-named system that evolves retrieval
algorithms \citep{nian2026rankevolve}.} an auto-research framework, answers
these challenges---and puts that
question to an objective test---with three design ideas.  \textbf{(1)~Executable Operating Protocols (EOPs).}  A workflow's phases,
dependencies, tools, confirmation gates, and branch/loop structure are
authored once as a versioned, semi-structured protocol and compiled to a
state machine the runtime itself enforces---keeping a long run on track
(challenge~iii) while the model performs each step
(Section~\ref{sec:agent}).  \textbf{(2)~A meta-meta-harness.}  Work nodes bind
complete, black-box coding-agent products (Claude Code, Codex, OpenHands)
through adapter contracts into review--repair and plan--merge flows: products
with different failure modes cross-check one another (challenge~ii) and
diversify proposals under a tight budget (challenge~i).  On our
incident-seeded oracle benchmark this lifts execution accuracy from the best
single-product $45.8\%$ to $62.5\%$ at matched budget
(Section~\ref{sec:execacc}), establishing \emph{execution accuracy} as a
binding constraint in this setting and showing that product composition can
improve it at matched budget.  \textbf{(3)~A knowledge layer.}
Findings---especially the lessons
of negative results---are recorded with full provenance and carried into
later iterations, so prior dead ends need not be silently re-derived
(challenge~i; Section~\ref{sec:knowledge}).

We validate end to end: a twelve-iteration deployment on the public HSTU
stack reported \ndcg{} 0.2192 against the published 0.2098 anchor on
MovieLens-20M LARGE and 0.1948 against 0.1895 on BASE, with first-class
negative results and logged leakage incidents (Section~\ref{sec:setup});
\execbench{}, built from those incidents---96 private tasks per repository,
locked snapshots, hidden oracles, a frozen inference-spend cap,
extended-budget/same-product-review/best-of-$N$ controls, a pre-specified
decorrelation analysis, and an equally controlled LitGPT
\citep{lightning2023litgpt} transfer split reported whatever its outcome
(Section~\ref{sec:execacc}); and a context-scope ablation isolating prompt
scope from enforcement (Section~\ref{sec:eop-ablation}).

The headline claim is pre-specified and gated on the primary HSTU corpus
against a frozen budget-matched baseline family, with the LitGPT split
reported whatever its outcome---a falsifiable protocol fixed before the
private runs, so the reported effects are confirmatory rather than
post-hoc.

\section{Anatomy of RankEvolve}
\label{sec:agent}
\re{} has three layers with deliberately different claims: an \emph{authoring
layer}, the semi-structured EOP document (Section~\ref{sec:eop}); a
\emph{control layer}, the \emph{meta-meta-harness} that compiles supported
markers to runtime state and dispatches work through adapter-bound
coding-agent products (Section~\ref{sec:metameta}); and a \emph{knowledge
layer} persisting hypotheses, patches, run manifests, evaluator provenance,
and promotion decisions (Section~\ref{sec:knowledge}).  This paper evaluates
the ML-evolution instance; transfer to arbitrary long-horizon domains is not
assumed.

\subsection{Executable Operating Protocols}
\label{sec:eop}
\re{}'s top-level abstraction is the \emph{Executable Operating Protocol}
(EOP): a versionable, semi-structured specification of a long-horizon
workflow, naming its \emph{phases}, \emph{dependencies}, required
\emph{tools}, \emph{user-confirmation gates}, and \emph{branch}/\emph{loop}
edges.  Rather than hand-coding---and debugging---a bespoke control flow for
every task, the workflow is declared once, so procedural knowledge outlives
brittle, throwaway scripts, in the spirit of MetaGPT's standardized
operating procedures \citep{hong2024metagpt}.  The document reads like an
operating procedure: structural markers are parsed and statically checked,
while a phase's natural-language body is delivered to its executor and is
not claimed to have formal semantics.  \re{}'s recommendation workflow is one such EOP
(Figure~\ref{fig:loop}; full text in Appendix~\ref{app:eop}): investigate,
research-and-propose, branch implement-experiment-and-analyze over selected
proposals, then summarize-and-evolve, looping while gains remain.  Gates are
first-class phases, so human control---proposal review and selection,
experiment configuration, iterate/stop---is explicit.  In the case study the
operator also authorized compute, vetoed one over-budget experiment, and gave
three high-level directions: favor new methods over stacks of add-ons, compare
like with like (BASE to BASE, LARGE to LARGE), and guarantee no test leakage
(Appendix~\ref{sec:discussion}).  Proposals are \emph{hypotheses}:
atomic, flag-gated changes combined into \emph{combos} and re-ranked not by
a numeric genetic operator but by an LLM critic over a persistent
leaderboard that promotes reference-beating combos to seed the next round.

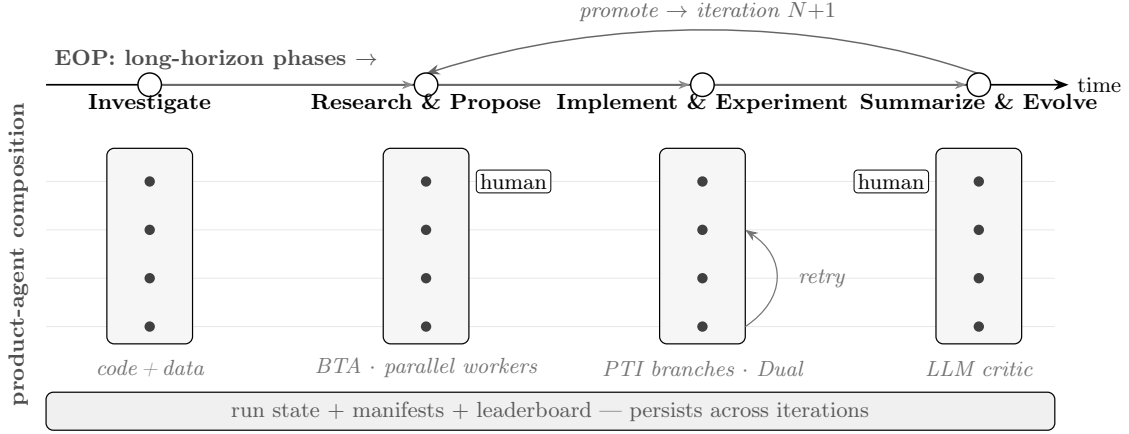
\begin{figure}[t!]
\centering
\resizebox{0.9\textwidth}{!}{%
\begin{tikzpicture}[
  >=Stealth,
  station/.style={draw, circle, minimum size=3.4mm, inner sep=0pt, fill=white, line width=0.6pt},
  pill/.style={draw, rounded corners=3pt, fill=gray!7, line width=0.5pt},
  dot/.style={circle, fill=black!75, minimum size=1.5mm, inner sep=0pt},
  pdot/.style={circle, fill=black!45, minimum size=1.0mm, inner sep=0pt},
  gate/.style={draw, rounded corners=1.5pt, fill=white, font=\small, inner sep=1.6pt},
  ann/.style={font=\small\itshape, text=black!55, align=center},
  dl/.style={font=\small, text=black!55, anchor=east}
]
% --- depth guide lines + left axis (harness) ---
\foreach \y in {-0.9,-1.6,-2.3,-3.0}{\draw[gray!18, line width=0.4pt] (0.9,\y)--(15.5,\y);}
\node[rotate=90, font=\small\bfseries, text=black!70] at (0.5,-1.95){product-agent composition};
% --- time axis (EOP horizon) ---
\draw[->, line width=0.7pt] (0.9,0.5)--(15.7,0.5) node[right, font=\small]{time};
\node[anchor=west, font=\small\bfseries, text=black!70] at (0.9,0.86){EOP: long-horizon phases $\rightarrow$};
% --- loop-back (iteration) ---
\draw[->, black!60, line width=0.6pt] (14.4,0.66) .. controls (12,1.5) and (9,1.5) .. (6.4,0.66)
  node[midway, above, font=\small\itshape, text=black!60]{promote $\rightarrow$ iteration $N{+}1$};
% --- P1 Investigate ---
\node[station] (s1) at (2.4,0.5){};
\node[font=\small\bfseries, anchor=south] at (2.4,-0.04){Investigate};
\draw[pill] (1.78,-0.42) rectangle (3.02,-3.25);
\node[dot] at (2.4,-0.9){}; \node[dot] at (2.4,-1.6){};
\node[dot] at (2.4,-2.3){}; \node[dot] at (2.4,-3.0){};
\node[ann] at (2.4,-3.6){code\,+\,data};
% --- P2 Research & Propose ---
\node[station] (s2) at (6.4,0.5){};
\node[font=\small\bfseries, anchor=south] at (6.4,-0.04){Research \& Propose};
\draw[pill] (5.78,-0.42) rectangle (7.02,-3.25);
\node[dot] at (6.4,-0.9){}; \node[dot] at (6.4,-1.6){};
\node[dot] at (6.4,-2.3){}; \node[dot] at (6.4,-3.0){};
\node[ann] at (6.4,-3.6){BTA $\cdot$ parallel workers};
\node[gate, anchor=west] at (7.12,-0.9){human};
% --- P3 Implement & Experiment ---
\node[station] (s3) at (10.4,0.5){};
\node[font=\small\bfseries, anchor=south] at (10.4,-0.04){Implement \& Experiment};
\draw[pill] (9.78,-0.42) rectangle (11.02,-3.25);
\node[dot] at (10.4,-0.9){}; \node[dot] at (10.4,-1.6){};
\node[dot] at (10.4,-2.3){}; \node[dot] at (10.4,-3.0){};
\draw[->, black!55, line width=0.5pt] (11.02,-3.0) .. controls (11.62,-2.6) and (11.62,-2.0) .. (11.02,-1.6);
\node[ann, anchor=west] at (11.66,-2.3){retry};
\node[ann] at (10.4,-3.6){PTI branches $\cdot$ Dual};
% --- P4 Evolve ---
\node[station] (s4) at (14.4,0.5){};
% \vphantom{y}: this label has no descender, so without it its south anchor
% would drop the baseline below the other three phase labels.
\node[font=\small\bfseries, anchor=south] at (14.4,-0.04){Summarize \& Evolve\vphantom{y}};
\draw[pill] (13.78,-0.42) rectangle (15.02,-3.25);
\node[dot] at (14.4,-0.9){}; \node[dot] at (14.4,-1.6){}; \node[dot] at (14.4,-2.3){}; \node[dot] at (14.4,-3.0){};
\node[ann] at (14.4,-3.6){LLM critic};
\node[gate, anchor=east] at (13.68,-0.9){human};
% --- artifact flow between phases ---
\draw[->, black!45] (s1)--(s2);
\draw[->, black!45] (s2)--(s3);
\draw[->, black!45] (s3)--(s4);
% --- persistent memory band ---
\draw[pill, fill=gray!12] (0.9,-3.95) rectangle (15.5,-4.5);
\node[font=\small, text=black!65] at (8.2,-4.225){run state $+$ manifests $+$ leaderboard --- persists across iterations};
\end{tikzpicture}}%
\caption{\textbf{Anatomy of \re{}: two orthogonal axes.}
\emph{Horizontally}, the EOP drives the long horizon---a compiled phase
sequence whose promoted results loop back for the next iteration
(Section~\ref{sec:eop}).  \emph{Vertically}, the meta-meta-harness supplies
depth and diversity: coding and review nodes bind complete products, while
routing, aggregation, and gates remain deterministic; a persistent
state/artifact store spans the run.  The diagram describes the deployed
integration boundary, not a novel graph formalism.}
\label{fig:loop}
\end{figure}

On the backend an EOP compiles to a \emph{state machine}: phases are states;
dependency, \texttt{branch}, join, and \texttt{goto} markers become
transitions; the runtime advances state and blocks at gates (branch fan-out
is a tested backend capability, not assumed from parsing).  The default
delivery policy injects the global preamble and current phase instruction,
not inactive phase bodies; StateFlow already motivates state-specific
prompting \citep{wu2024stateflow}, so whether this policy improves
\emph{final code correctness} at fixed runtime is tested in
Section~\ref{sec:eop-ablation}, not asserted.

\paragraph{Runtime contract.}
For protocol $P$, compilation produces a graph $C(P)$ and runtime state
$\sigma=(q,V,A,G,B,R,T)$---active phase, protocol variables, committed
artifacts, gate status, live branches, attempt/budget counters, append-only
event trace---advanced by typed events
($\delta(\sigma,e)\rightarrow\sigma'$: \textsc{node-complete},
\textsc{gate-approved}, \textsc{branch-failed},
\textsc{budget-exhausted}).  The enforcement contract
(Appendix~\ref{app:protocol-details}) makes the guarantee boundary explicit:
dependencies, gates, branch/join cardinality, loop budgets, tool allowlists, and
checkpoint/restart are checked statically or enforced at runtime, while
semantic code correctness is expressly \emph{not} guaranteed.

A historical pilot motivates, but does not validate, this design: ten runs
received the whole procedure as one in-context playbook, and none was fully
faithful---all ten ran only a subset of selected proposals, four stopped
early, four skipped a gate, three called tools off-spec (audit table in
Appendix~\ref{app:protocol-details}; Figure~\ref{fig:faithfulness} shows the
accumulation).  The pilot lacks a matched EOP-runtime arm and changes control
and context delivery together, so it supports only the existence of a failure
mode; the controlled ablation in Section~\ref{sec:eop-ablation} is the
required test.

\begin{figure}[t]
\centering
\begin{tikzpicture}[x=1.45cm, y=0.36cm,
  ln/.style={line width=0.9pt},
  lg/.style={font=\small, anchor=west, inner sep=0pt}]
% --- axes ---
\draw[gray!25, line width=0.4pt] (0,2)--(3.15,2); \draw[gray!25, line width=0.4pt] (0,4)--(3.15,4);
\draw[gray!25, line width=0.4pt] (0,6)--(3.15,6); \draw[gray!25, line width=0.4pt] (0,8)--(3.15,8);
\draw[line width=0.6pt] (0,0)--(3.15,0);
\draw[line width=0.6pt] (0,0)--(0,8.4);
\foreach \y in {0,2,4,6,8} \node[font=\small, anchor=east, inner sep=2pt] at (0,\y) {\y};
\foreach \x/\l in {0.55/1, 1.6/2, 2.65/3} {\draw[line width=0.6pt] (\x,0)--(\x,-0.25);
  \node[font=\small, anchor=north, inner sep=3pt] at (\x,-0.2) {\l};}
\node[font=\small, anchor=north] at (1.6,-1.5) {Evolution iteration};
\node[font=\small, rotate=90, anchor=south] at (-0.42,4) {Runs with violation};
% --- series ---
\draw[ln, red] (0.55,3)--(1.6,7)--(2.65,7);
\foreach \x/\y in {0.55/3, 1.6/7, 2.65/7} \fill[red] (\x,\y) circle (2.1pt);
\draw[ln, blue] (0.55,0)--(1.6,3)--(2.65,1);
\foreach \x/\y in {0.55/0, 1.6/3, 2.65/1} \fill[blue] ($(\x,\y)+(-2pt,-2pt)$) rectangle ++(4pt,4pt);
\draw[ln, green!55!black] (0.55,1)--(1.6,2)--(2.65,3);
\foreach \x/\y in {0.55/1, 1.6/2, 2.65/3} \fill[green!55!black] ($(\x,\y)+(-2.4pt,-1.8pt)$)--($(\x,\y)+(2.4pt,-1.8pt)$)--($(\x,\y)+(0,2.6pt)$)--cycle;
\draw[ln, orange] (0.55,0)--(1.6,2)--(2.65,2);
\foreach \x/\y in {0.55/0, 1.6/2, 2.65/2} \fill[orange] ($(\x,\y)+(0,2.6pt)$)--($(\x,\y)+(2.4pt,0)$)--($(\x,\y)+(0,-2.6pt)$)--($(\x,\y)+(-2.4pt,0)$)--cycle;
\draw[ln, violet] (0.55,0)--(1.6,2)--(2.65,3);
\foreach \x/\y in {0.55/0, 1.6/2, 2.65/3} {\draw[violet, line width=0.8pt] (\x,\y) circle (2.1pt);}
\draw[ln, brown, dashed] (0.55,2)--(1.6,0)--(2.65,0);
\foreach \x/\y in {0.55/2, 1.6/0, 2.65/0} {\draw[brown, line width=0.8pt] ($(\x,\y)+(-2pt,-2pt)$)--($(\x,\y)+(2pt,2pt)$) ($(\x,\y)+(-2pt,2pt)$)--($(\x,\y)+(2pt,-2pt)$);}
\draw[ln, black, densely dotted] (0.55,0)--(1.6,1)--(2.65,2);
\foreach \x/\y in {0.55/0, 1.6/1, 2.65/2} {\draw[black, line width=0.8pt] ($(\x,\y)+(-2pt,0)$)--($(\x,\y)+(2pt,0)$) ($(\x,\y)+(0,-2.2pt)$)--($(\x,\y)+(0,2.2pt)$);}
% --- legend (two columns, 9pt) ---
\begin{scope}[shift={(0,-3.1)}]
\draw[ln, red] (0,0)--(0.13,0); \fill[red] (0.065,0) circle (2.1pt);
\node[lg] at (0.17,0) {Partial branch};
\draw[ln, blue] (0,-1.15)--(0.13,-1.15); \fill[blue] ($(0.065,-1.15)+(-2pt,-2pt)$) rectangle ++(4pt,4pt);
\node[lg] at (0.17,-1.15) {Premature stop};
\draw[ln, green!55!black] (0,-2.3)--(0.13,-2.3);
\fill[green!55!black] ($(0.065,-2.3)+(-2.4pt,-1.8pt)$)--($(0.065,-2.3)+(2.4pt,-1.8pt)$)--($(0.065,-2.3)+(0,2.6pt)$)--cycle;
\node[lg] at (0.17,-2.3) {Skipped gate};
\draw[ln, orange] (0,-3.45)--(0.13,-3.45);
\fill[orange] ($(0.065,-3.45)+(0,2.6pt)$)--($(0.065,-3.45)+(2.4pt,0)$)--($(0.065,-3.45)+(0,-2.6pt)$)--($(0.065,-3.45)+(-2.4pt,0)$)--cycle;
\node[lg] at (0.17,-3.45) {Off-spec tool args};
\draw[ln, violet] (1.62,0)--(1.75,0); \draw[violet, line width=0.8pt] (1.685,0) circle (2.1pt);
\node[lg] at (1.79,0) {Redundant re-work};
\draw[ln, brown, dashed] (1.62,-1.15)--(1.75,-1.15);
\draw[brown, line width=0.8pt] ($(1.685,-1.15)+(-2pt,-2pt)$)--($(1.685,-1.15)+(2pt,2pt)$) ($(1.685,-1.15)+(-2pt,2pt)$)--($(1.685,-1.15)+(2pt,-2pt)$);
\node[lg] at (1.79,-1.15) {Merged phases};
\draw[ln, black, densely dotted] (1.62,-2.3)--(1.75,-2.3);
\draw[black, line width=0.8pt] ($(1.685,-2.3)+(-2pt,0)$)--($(1.685,-2.3)+(2pt,0)$) ($(1.685,-2.3)+(0,-2.2pt)$)--($(1.685,-2.3)+(0,2.2pt)$);
\node[lg] at (1.79,-2.3) {Other deviation};
\end{scope}
\end{tikzpicture}
\caption{Per-iteration process-faithfulness violations across the ten
full-context playbook runs (no runtime enforcement).  Deviations are rare in
iteration~1 and proliferate from iteration~2 as accumulated context crowds
out the current-step instruction.  Runs reaching each iteration:
$n{=}10,10,7$; a point counts runs exhibiting that violation \emph{at} that
iteration, so a run may contribute to several (per-run totals in
Table~\ref{tab:faithfulness}).  The pilot lacks a runtime-controlled arm, so
it motivates rather than measures the EOP's effect; the controlled test is
Section~\ref{sec:eop-ablation}.}
\label{fig:faithfulness}
\end{figure}
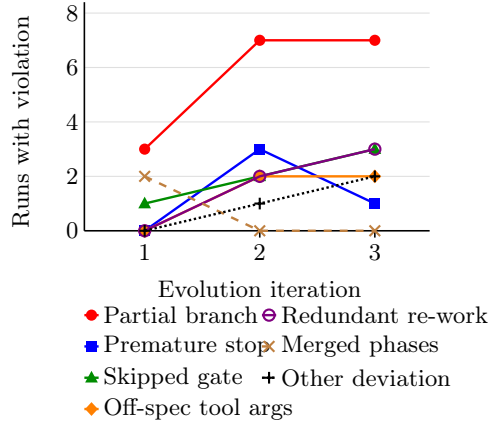

\subsection{A meta-meta-harness}
\label{sec:metameta}

\re{}'s control layer is a \emph{meta-meta-harness}: each commercial coding
product is itself a harness over a raw model---planner, context management,
tools, and sandbox---and \re{} harnesses these harnesses in turn, standing
two harnessing levels above the model and composing complete products as the
work nodes of its execution graph rather than issuing raw model calls.
Agents-as-graph-nodes is an established
pattern \citep{zhuge2024gptswarm,langchain2026langgraph}; the distinctive
choices here are the node's granularity and opacity---a complete
\emph{product} bound as a black box---and a runtime that keeps control flow
outside the products.

The binding is an adapter contract: it receives the task, repository
snapshot, allowed tools, budget, and prior artifacts, and returns a patch,
terminal status, event trace, and usage record; it owns sandboxing,
non-interactive invocation, timeout, and usage collection---not the product's
internal planning.  This makes Claude Code \citep{anthropic2026claudecode},
Codex \citep{openai2026codex}, and OpenHands \citep{wang2025openhands}
substitutable at \emph{coding or review} nodes; routing, aggregation, gates,
and metrics remain deterministic runtime functions or narrow model calls, so
we do not claim every graph node is itself an agent.

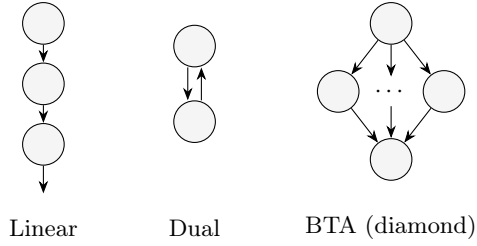
\begin{figure}[t]
\centering
\begin{tikzpicture}[>=Stealth,
  n/.style={draw, circle, fill=gray!8, minimum size=5.5mm, inner sep=0pt}]
% --- linear ---
\node[n] (l1) at (0,0) {};
\node[n] (l2) at (0,-0.8) {};
\node[n] (l3) at (0,-1.6) {};
\draw[->] (l1)--(l2); \draw[->] (l2)--(l3); \draw[->] (l3)--(0,-2.25);
\node[font=\small] at (0,-2.7) {Linear};
% --- dual ---
\node[n] (d1) at (2,-0.3) {};
\node[n] (d2) at (2,-1.3) {};
\draw[->,transform canvas={xshift=-0.9mm}] (d1)--(d2);
\draw[->,transform canvas={xshift=0.9mm}] (d2)--(d1);
\node[font=\small] at (2,-2.7) {Dual};
% --- diamond ---
\node[n] (t)  at (4.6,0) {};
\node[n] (ml) at (3.9,-0.9) {};
\node[font=\small] (mc) at (4.6,-0.9) {$\cdots$};
\node[n] (mr) at (5.3,-0.9) {};
\node[n] (bb) at (4.6,-1.8) {};
\draw[->] (t)--(ml); \draw[->] (t)--(mc); \draw[->] (t)--(mr);
\draw[->] (ml)--(bb); \draw[->] (mc)--(bb); \draw[->] (mr)--(bb);
\node[font=\small] at (4.6,-2.7) {BTA (diamond)};
\end{tikzpicture}
\caption{Execution patterns instantiated by the runtime.  Circles denote work
nodes, which may bind to a coding-agent product, a deterministic function, or
a narrow model call.  Disabling the diamond's aggregator yields a tree;
disabling its breakdown yields a chain.  These topologies are standard; our object of study is
the product-agent adapter boundary and its measured reliability.}
\label{fig:topologies}
\end{figure}

The runtime instantiates familiar execution patterns
(Figure~\ref{fig:topologies}): \textbf{Linear}, a chain with loop-back;
\textbf{Dual}, a propose$\rightarrow$review$\rightarrow$fix consensus
loop---in effect, internal peer review---that repeats until the reviewer
approves or residual severity falls below a threshold, within bounded rounds;
and \textbf{Breakdown-then-Aggregate (BTA)}, a diamond splitting a task into
bounded-concurrency sub-tasks and synthesizing results, the fan-out behind
the parallel research workers of Figure~\ref{fig:loop}.  These compose into
flows such as \textbf{Plan-then-Implement (PTI)}\@.
The recovery contract is specific: state and artifacts checkpoint at node
boundaries; transient invocations get a bounded retry budget; a timed-out
node fails closed; a resumed run starts from the last committed boundary; and
recovery claims are limited to the contract's tested paths
(Appendix~\ref{app:protocol-details}).  The deployed flow is human-authored
YAML (plan stage in Appendix~\ref{app:yaml})---a \emph{plan-merge} composition: parallel
planning cross-merged to convergence, then dual implement--review
(Figure~\ref{fig:winflow}); automatic workflow search is outside our
contribution, being studied directly by GPTSwarm, ADAS, and AFlow
\citep{zhuge2024gptswarm,hu2025adas,zhang2025aflow}.

Type-level interfaces are in Appendix~\ref{app:primitives}.  These patterns
form the \emph{vertical} axis of Figure~\ref{fig:loop}---each phase's depth
and diversity---while the EOP fixes the horizontal order; whether
heterogeneous cross-checking improves correctness at equal budget is the
hypothesis of Section~\ref{sec:execacc}, not an assumption.

\subsection{The knowledge layer: artifact and evidence state}
\label{sec:knowledge}

A long-horizon agent that forgets re-derives the same dead ends, so a long
run needs durable evidence.  \re{} therefore implements a \emph{knowledge
layer} with two linked responsibilities.  First, its artifact/evidence store
records, per proposal, repository/environment hashes, prompt, EOP and product
versions, patch, test outputs, training configuration, split, checkpoints,
metrics, cost, and the promotion decision, with negative outcomes as
first-class leaderboard rows---supporting resumption and claim-to-run
auditing.  Second, its experiential memory writes self-contained notes
\emph{beside the code they describe}, abstracts transferable lessons into a
central wiki that cites those notes as evidence, and links both tiers through
an entity graph.  A reconciliation agent keeps the wiki the single source of
truth, while a per-phase read path injects only the few relevant lessons,
preserving the EOP's per-step context discipline
(Section~\ref{sec:eop}).  This implemented layer was active in the reported
twelve-iteration HSTU deployment: it persisted negative results and incident
lessons and made them available to later phases.  It is part of \re{}'s
system contribution and is detailed against prior agent-memory mechanisms in
Appendix~\ref{app:knowledge}.  The study evaluates the layer as part of the
end-to-end system but does not include a memory-on/off ablation, so it does
not isolate the layer's marginal effect on proposal quality
\citep{shinn2023reflexion,zhao2024expel}.

\section{HSTU Deployment Case Study}
\label{sec:setup}

\subsection{Scope, target, and evaluation}
\label{sec:hstu}

Our target is HSTU \citep{zhai2024hstu}, a generative recommender; we use its
\emph{public} implementation and public MovieLens data
\citep{harper2015movielens}, so all results are at public-benchmark scale.  ML-20M is mature and heavily benchmarked---published methods cluster in a
narrow band---so consistent \ndcg{} gains are hard-won: a deliberately
stringent test.  HSTU replaces the Transformer block with
SiLU-normalized attention, a fused UVQK projection, and a learnable relative
bucketed time-and-position bias---the last is why explicit time-decay add-ons
prove redundant (Appendix~\ref{sec:appendix-hstu}).  BASE/LARGE configurations are tabulated in
Appendix~\ref{sec:appendix-repro}.  Published MovieLens-20M anchors are 0.1895
(BASE) and 0.2098 (LARGE).

\paragraph{Evaluation discipline.}\label{sec:eval}
We report \emph{canonical full-test} \ndcg{} (leave-one-out, full-corpus
ranking over all held-out users), distinct from a faster \emph{subset}
evaluation that ran ${\approx}0.005$--$0.010$ higher in the logged runs;
checkpoint evaluations from one run are not independent seeds, so ``best
checkpoint'' values are descriptive history (Table~\ref{tab:main}).  This separation was the study's most important
methodological discipline: without it, the system repeatedly chased phantom
0.005-scale ``gains'' from subset and checkpoint noise.

\paragraph{Compute.}\label{sec:compute}
The workspace contains ${\sim}55$ leaderboard entries across 49 experiment
directories on 1--8 NVIDIA H200s; we report run counts rather than an
unverifiable GPU-hour aggregate.

\begin{table}[t]
\centering
\small
\setlength{\tabcolsep}{4pt}
\begin{tabular}{@{}c>{\raggedright\arraybackslash}p{4.6cm}>{\raggedright\arraybackslash}p{2.3cm}>{\raggedright\arraybackslash}p{7.8cm}@{}}
\toprule
\textbf{Iter.} & \textbf{Intervention} & \textbf{Outcome} &
\textbf{Finding and claim status} \\
\midrule
0 & Published HSTU anchor & 0.2098 & Published comparison point \\
\midrule
\multicolumn{4}{@{}l}{\emph{Stage~I --- compose known components (\textsc{SYNAPSE}); one objective bet}} \\
1--2 & SSD-style input compression (IC) $+$ PRISM user conditioning & 0.2140 &
$+2.0\%$; measured historical endpoint \\
3 & $+$ multi-token scoring head & \textbf{0.2161} &
\textsc{SYNAPSE}, $+3.0\%$: two pooled facets (recent intent, long-term
taste), one ANN call; measured historical endpoint \\
4 & Preference optimization (DPO/IPO/SimPO) & DPO: $-35.9\%$ vs.\ its 0.2191
reference &
All three used target-derived negatives (pre-fix mining) and hurt as
preference accuracy approached 1; reward over-optimization remains a hypothesis.
A leak-free re-derivation mining only internal positions is neutral
($\Delta{\approx}0$). Rejected \\
\midrule
\multicolumn{4}{@{}l}{\emph{Stage~II --- refine (diminishing returns)}} \\
5 & FLUID elapsed-gap time decay & ties 0.2161 &
Redundant with HSTU's learnable relative time bias \\
6 & Failure-overlap diagnostic & 75.2\% overlap &
On a 0.2127 checkpoint of the 0.2140 stack, last-position failures persist
one position earlier; descriptive, not causal \\
7 & Focal / tail reweighting & no lift &
Three implementation bugs fixed, then nine runs without gain. Rejected \\
8 & Deep failure analysis; multi-position training augmentation (MPTA) &
0.2154 & The model's top-ranked non-targets share ${\approx}0.236$ genre
Jaccard with the target (``genre right, item wrong''); MPTA adds $+0.7\%$ to
the 0.2140 stack it extends, within checkpoint noise, and sets no new best \\
\midrule
\multicolumn{4}{@{}l}{\emph{Stage~III --- push the ceiling}} \\
9 & Additive genre side-feature (\textsc{UDK}) & \textbf{0.2192} &
Largest reported ML-20M LARGE endpoint, $+4.48\%$, fine-tuned from the
0.2140 stack (not \textsc{SYNAPSE}) once a v1 feature leak was fixed; a gain
from added metadata, not architectural novelty \\
10 & Position-averaged test-time augmentation & ${\approx}0.2190$ &
No gain: each position has a different target, so there is no shared label
to average. Limited to this estimator. Rejected \\
11 & \textsc{UDK} channels: genre / year / popularity & ${\approx}0.2192$ &
All channels saturate at the same ${\approx}0.219$ ceiling \\
12 & Boost-last-$K$ at inference; training-side variant and new methods &
no gain (inference) & Inference-time variants over $K$ and boost strength stay
within $\pm0.0003$ of 0.2192, inside checkpoint noise, so none counts as a new
endpoint; the training-side variant and the new methods did not finish
within the study and are not reported \\
\bottomrule
\end{tabular}
\caption{Twelve-iteration case history on ML-20M\,$\times$\,LARGE\@.  An
iteration is one or more propose$\rightarrow$implement$\rightarrow$evaluate
turns; iterations are grouped by theme into three stages and numbered by
stage, not strictly in run order.  \ndcg{} values are \emph{reported historical
endpoints} (best checkpoints, not independent seeds); the final column scopes
each claim.  Bold marks the stage milestones.  Relative gains are over the
published anchor (0.2098; 0.1895 for the BASE result below) unless a row names
another reference (MPTA: the 0.2140 stack; DPO: its own reference,
Appendix~\ref{sec:negatives}).  DPO's 0.2191 reference is not a Stage~I
endpoint; iteration~4 is placed in Stage~I by theme.
Background recipe sweeps are not listed; among them, the one clean BASE win,
dropout $0.2\rightarrow0.1$ in Stage~II, gave 0.1948 ($+2.80\%$;
Table~\ref{tab:main}), and a $500\rightarrow700$ context extension lost
accuracy (Appendix~\ref{sec:negatives}).}
\label{tab:iters}
\end{table}

\subsection{Three-stage trajectory}
\label{sec:trajectory}

The twelve iterations summarize as three \emph{stages}
(Table~\ref{tab:iters}, whose final column scopes each claim): Stage~I
composed known architectural components to 0.2161; Stage~II's refinements
were mostly within noise; Stage~III's genre side feature reached the largest
endpoint, 0.2192; and the preference-optimization direction produced the
largest regression.  These illustrate the workload and its failure modes,
not a claim of architectural novelty.

\subsection{Reported endpoints and integrity boundary}
\label{sec:results}

Table~\ref{tab:main} separates published anchors from internal
reproductions: MovieLens-32M has no published HSTU anchor, so those rows
cannot support a ``published-baseline'' statement, and the 0.2192 endpoint
comes from an added genre side-feature, not a new modeling primitive.
Full diagnostic and negative-result narratives are in
Appendix~\ref{sec:appendix-hstu}.

\begin{table}[t]
\centering
\small
\begin{tabular}{@{}lccc@{}}
\toprule
\textbf{Dimension} & \textbf{Reported} & \textbf{Reference} &
\textbf{Relative $\Delta$} \\
\midrule
ML-20M $\times$ BASE  & 0.1948 & 0.1895$^{p}$ & $+2.80\%$ \\
ML-20M $\times$ LARGE & 0.2192 & 0.2098$^{p}$ & $+4.48\%$ \\
ML-32M $\times$ BASE  & 0.1562 & 0.1488$^{i}$ & $+4.97\%$ \\
ML-32M $\times$ LARGE & 0.1726 & 0.1660$^{i}$ & $+3.98\%$ \\
\bottomrule
\end{tabular}
\caption{Reported best-checkpoint endpoints from the case history.  The
ML-32M BASE endpoint is a five-component stack
(PRISM${+}$IC${+}$time-decay${+}$dropout$=$0.1${+}$hard negatives) and the
ML-32M LARGE endpoint a PRISM${+}$IC${+}$time-decay stack; neither appears in
Table~\ref{tab:iters}, which covers ML-20M only.  $^{p}$Published HSTU anchor; $^{i}$internal reproduction, not a published
baseline.}
\label{tab:main}
\end{table}

\paragraph{Cross-dataset transfer.}
To check the discovered levers are portable, the deployment's two exported
recipes were replayed on four further sequential-recommendation datasets with
no per-dataset re-tuning: Stage~I \textsc{SYNAPSE}
(Appendix~\ref{sec:appendix-traj}), then the Stage~III genre side-feature
\textsc{UDK} on top (on ML-20M, \textsc{UDK} was instead fine-tuned from the
0.2140 stack; Table~\ref{tab:iters}).  Both improve over vanilla HSTU on all
four (Table~\ref{tab:generalize}), cumulatively by up to $+25\%$, though
\textsc{UDK} adds little on Foursquare-TKY ($0.0200\rightarrow0.0202$).  The
beyond-recommenders question belongs to the LitGPT split
(Section~\ref{sec:execacc}).

\begin{table}[t]
\centering
\small
\setlength{\tabcolsep}{3pt}
\begin{tabular}{@{}lccc@{}}
\toprule
\textbf{Dataset} & \textbf{Vanilla} & \textbf{$+$\textsc{SYNAPSE}} &
\textbf{${+}{+}$\textsc{UDK}} \\
\midrule
Foursquare-TKY & 0.0181 & 0.0200~{\small($+10\%$)} & 0.0202~{\small($+12\%$)} \\
Foursquare-NYC & 0.0156 & 0.0180~{\small($+15\%$)} & 0.0195~{\small($+25\%$)} \\
Gowalla        & 0.0441 & 0.0460~{\small($+4.3\%$)} & 0.0467~{\small($+5.9\%$)} \\
Yelp           & 0.0321 & 0.0359~{\small($+11.8\%$)} & 0.0382~{\small($+19.0\%$)} \\
\bottomrule
\end{tabular}
\caption{Cross-dataset transfer of the deployment's discovered recipes
(\ndcg{}, canonical full-test, one run per cell; columns are cumulative---${+}{+}$\textsc{UDK}
adds the Stage~III side-feature on top of $+$\textsc{SYNAPSE}; percentages over
vanilla).}
\label{tab:generalize}
\end{table}

\section{Objective Execution-Accuracy Evaluation}
\label{sec:execacc}

Why compose products at all, rather than call one strong coding agent?  On
mature ML code even a single maximum-effort plan-then-execute pass can
produce plans with critical defects---reinventing facilities the codebase
already provides, calling deprecated APIs that would fail CI, writing to
wrong paths, or omitting build targets---and one bad patch can corrupt an
iteration or, worse, silently inflate a metric.
\execbench{} turns the deployment's logged incidents into an objective,
budget-controlled benchmark over two coding-agent products, Claude Code (CC)
and Codex.  It asks whether product composition improves
executable correctness (\textbf{RQ1}); whether any gain survives an equal
observable budget on HSTU (\textbf{RQ2}), with a pre-specified LitGPT
transfer version; whether error complementarity predicts realized
review--repair gain (\textbf{RQ3}); and whether current-step context helps at
fixed enforcement (\textbf{RQ4}).  The primary pre-specified hypothesis is
that CC$\rightarrow$Codex$\rightarrow$CC exceeds $\max_{m\in\mathcal{B}}
\mathrm{EA}_m$ on the private HSTU split, where $\mathcal{B}$ is the frozen
matched-budget baseline family; we use the phrase ``heterogeneous
composition buys execution accuracy'' only if the task-clustered 95\%
interval excludes zero \emph{and} the effect exceeds a smallest effect of
interest frozen after development-only power analysis.  The same contrast on
LitGPT is reported with its interval whatever its outcome---extending the
claim beyond recommenders if positive, bounding it to the deployed domain if
null---and is published either way, not suppressed.

\paragraph{Corpus, oracle, and conditions.}
Each repository contributes 96 private tasks: the public HSTU codebase,
seeded by the case study's logged incidents, and LitGPT, a non-recommender
training codebase \citep{lightning2023litgpt}.  Tasks balance six incident
families---data provenance and leakage, tensor routing, gradient flow,
train/eval mode, metric semantics, and configuration wiring---half new
implementation, half repair.  Every item pairs an immutable snapshot with a
hidden executable oracle that passes only when a patch \emph{simultaneously}
satisfies the regression suite, task-specific behavioral checks,
scientific-safety invariants, and evaluator-integrity checks; the primary
endpoint is all-oracle execution accuracy (EA), and we separately report the
\emph{silent critical-defect rate} (CDR): an oracle-confirmed fault that
leaves the patch runnable but can invalidate a scientific conclusion
(leakage, dead feature paths, broken gradients, wrong evaluation semantics).
CC runs Claude Opus~4.8 and Codex runs GPT-5.6, each at its maximum
reasoning-effort setting, in every condition.  Six conditions run from a
fresh sandbox with identical snapshot, tools,
network policy, prompt, and timeout, in randomized order: one-pass CC; one
continuous CC session at the full composition budget; independent CC
candidates with oracle-blind selection/repair;
CC$\rightarrow$CC$\rightarrow$CC implement--review--repair;
CC$\rightarrow$Codex$\rightarrow$CC; and the reversed
Codex$\rightarrow$CC$\rightarrow$Codex.  An oracle-selected best-of-$N$ is
computed as a non-deployable upper bound
(Appendix~\ref{app:protocol-details}).  A resource envelope $B$, frozen
before the private evaluation, gives every flow the same three roles,
per-node limits, tool permissions, action caps, and wall-clock and spend
caps; timeouts and overruns score as failures and stay in denominators.  Proprietary products do not expose provider-side FLOPs, so
``matched observable inference budget'' is the accurate term---not equality
of hidden hardware compute.  Pass/fail contrasts use exact McNemar or paired randomization, effect sizes
use task-clustered bootstrap intervals, and the primary test is the single
max-over-$\mathcal{B}$ contrast with Holm-corrected secondaries.  Oracle
formalism, per-family counts, curation and leakage controls, and full decision
rules are in Appendix~\ref{app:protocol-details}.

\begin{table}[t]
\centering
\small
\setlength{\tabcolsep}{4pt}
\begin{tabular}{@{}>{\raggedright\arraybackslash}p{4.5cm}cccccc@{}}
\toprule
\textbf{Condition} & \textbf{HSTU EA} $\uparrow$ &
\textbf{HSTU CDR} $\downarrow$ & \textbf{LitGPT EA} $\uparrow$ &
\textbf{LitGPT CDR} $\downarrow$ & \textbf{Cost/task} $\downarrow$ &
\textbf{Tool actions} $\downarrow$ \\
\midrule
CC, one pass & $22.9${\small$\,\pm10.5$} & $35.4${\small$\,\pm12.0$} & $20.8${\small$\,\pm10.1$} & $37.5${\small$\,\pm12.1$} &
$0.41$ & $37$ \\
CC, extended budget $B$ & $33.3${\small$\,\pm11.8$} & $27.1${\small$\,\pm11.1$} & $29.2${\small$\,\pm11.4$} &
$29.2${\small$\,\pm11.4$} & $0.98$ & $94$ \\
CC best-of-$N$ + blind selection & $43.8${\small$\,\pm12.4$} & $18.8${\small$\,\pm9.8$} &
$39.6${\small$\,\pm12.2$} & $20.8${\small$\,\pm10.1$} & $0.95$ & $90$ \\
CC$\rightarrow$CC$\rightarrow$CC & $45.8${\small$\,\pm12.5$} & $16.7${\small$\,\pm9.3$} &
$43.8${\small$\,\pm12.4$} & $18.8${\small$\,\pm9.8$} & $0.97$ & $103$ \\
\textbf{CC$\rightarrow$Codex$\rightarrow$CC} & $\mathbf{62.5}${\small$\,\pm12.1$} & $\mathbf{10.4}${\small$\,\pm7.6$} &
$\mathbf{56.2}${\small$\,\pm12.4$} & $\mathbf{12.5}${\small$\,\pm8.3$} & $1.02$ & $109$ \\
Codex$\rightarrow$CC$\rightarrow$Codex & $56.2${\small$\,\pm12.4$} & $12.5${\small$\,\pm8.3$} &
$50.0${\small$\,\pm12.5$} & $14.6${\small$\,\pm8.8$} & $1.00$ & $106$ \\
\midrule
Plan-merge flow (CC$\,\parallel\,$Codex), 2nd stage$^{\dagger}$ & $70.8${\small$\,\pm11.4$} & $6.2${\small$\,\pm6.0$} &
$64.6${\small$\,\pm12.0$} & $8.3${\small$\,\pm6.9$} & $2.40$ & $250$ \\
\bottomrule
\end{tabular}
\caption{Primary execution accuracy by condition ($n{=}96$ tasks
per repository); stage-1 rows 2--6 are resource-matched to envelope $B$,
with one-pass CC as an unmatched cheap reference.  HSTU columns
carry the primary claim, LitGPT columns the pre-specified transfer
contrast.  Cells: task-level mean \% with 95\% task-clustered intervals;
cost USD/task; tool actions per-task medians.  Bold: best stage-1 condition
on both EA and CDR at parity of budget.  The rule-separated final row is the
deployed plan-merge flow, specified after stage-1 unblinding and excluded
from the stage-1 contrast and Holm family; $^{\dagger}$it runs at
\emph{natural} (uncapped) cost, unlike the stage-1 rows
(Section~\ref{sec:stage2}).}
\label{tab:execacc}
\end{table}

\subsection{Results}
\label{sec:execacc-results}

Table~\ref{tab:execacc} reports the frozen private evaluation.
\textbf{RQ1/RQ2.}  On HSTU the heterogeneous pair reaches $62.5\%$ EA, above
every budget-matched baseline (one-pass $22.9$, extended-budget $33.3$, blind
best-of-$N$ $43.8$, CC$\rightarrow$CC$\rightarrow$CC $45.8$); the paired
contrast against the strongest baseline is $+16.7$ points (95\%
cluster-robust CI $[6.6,26.7]$, $p{<}0.001$; twenty discordant tasks,
eighteen rescued vs.\ two broken), clearing zero and the pre-specified
smallest effect: heterogeneous composition buys execution accuracy on the
deployed domain.  CDR falls in
step, $35.4\%$ to $10.4\%$---the lowest of the budget-matched
conditions---so the gain is fewer silent faults, not CI churn.  Extra same-product budget helps
($+10.4$ extended; $+12.5$ homogeneous review) but plateaus below the pair:
buying \emph{different} products, not more of the same, delivers the largest
increment.  The reversed pair also beats all homogeneous conditions
($56.2\%$) but trails forward by $6.2$ points---which product implements
versus reviews matters.  \textbf{Transfer.}  On LitGPT the forward
pair leads $56.2\%$ vs.\ $43.8\%$, a paired $+12.5$ (95\% CI $[3.0,22.0]$,
$p{=}0.008$): reported as specified in advance and regardless of outcome, the effect
\emph{replicates} on a non-recommender codebase, extending the confirmatory
claim beyond the deployed domain.  The five envelope-$B$ conditions run at
$\$0.95$--$\$1.02$/task and $90$--$109$ actions (one-pass CC is the
deliberately unmatched cheap reference), so the improvement is not bought
with budget.

\subsection{Second stage: the deployed plan-merge topology}
\label{sec:stage2}

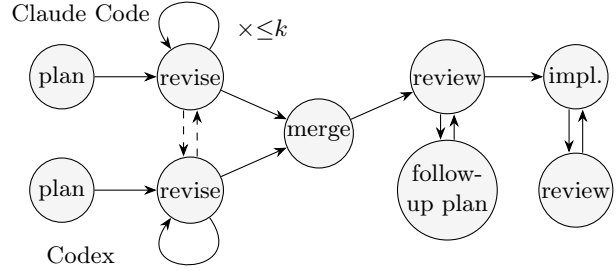
\begin{figure}[t]
\centering
\begin{tikzpicture}[>=Stealth,
  n/.style={draw, circle, align=center, font=\small, minimum size=8.5mm,
    inner sep=0.5pt, fill=gray!8}]
\node[n] (p1) at (0.7,0) {plan};
\node[n] (p2) at (0.7,-1.5) {plan};
\node[n] (r1) at (2.4,0) {revise};
\node[n] (r2) at (2.4,-1.5) {revise};
\node[n] (m)  at (4.1,-0.75) {merge};
\node[n] (rv) at (5.8,0) {review};
\node[n] (fp) at (5.8,-1.5) {follow-\\up plan};
\node[n] (im) at (7.5,0) {impl.};
\node[n] (rw) at (7.5,-1.5) {review};
\node[font=\small, anchor=south] at (0.95,0.62) {Claude Code};
\node[font=\small, anchor=north] at (0.95,-2.12) {Codex};
\draw[->] (p1)--(r1); \draw[->] (p2)--(r2);
\draw[->, dashed, transform canvas={xshift=-0.9mm}] (r1)--(r2);
\draw[->, dashed, transform canvas={xshift=0.9mm}] (r2)--(r1);
\draw[->] (r1) to[out=55,in=125,looseness=5] (r1);
\draw[->] (r2) to[out=-55,in=-125,looseness=5] (r2);
\node[font=\small] at (3.35,0.62) {$\times{\le}k$};
\draw[->] (r1)--(m); \draw[->] (r2)--(m);
\draw[->] (m)--(rv);
\draw[->, transform canvas={xshift=-0.9mm}] (rv)--(fp);
\draw[->, transform canvas={xshift=0.9mm}] (fp)--(rv);
\draw[->] (rv)--(im);
\draw[->, transform canvas={xshift=-0.9mm}] (im)--(rw);
\draw[->, transform canvas={xshift=0.9mm}] (rw)--(im);
\end{tikzpicture}%
\caption{The deployed plan-merge topology (second-stage condition), one
planning lane bound to Claude Code and the other to Codex: each lane drafts
and iteratively revises its plan while reading
the other lane's latest draft (dashed; ${\le}k$ rounds, per-lane stop
votes), a single aggregation integrates a winner (\emph{merge}), and a
review--follow-up-plan dual then an implement--review dual finish.}
\label{fig:winflow}
\end{figure}

After stage-1 unblinding---every stage-1 inference already complete---we
specified one further condition: the deployed \emph{plan-merge} flow
(Figure~\ref{fig:winflow}), with its CC$\,\parallel\,$Codex planning lanes
and the same product settings as stage~1.  The topology is the deployment's
own configuration, fixed before stage~1 (plan stage listed in
Appendix~\ref{app:yaml}); only its budget partitioning is adapted
to envelope~$B$ (Appendix~\ref{app:protocol-details}).  Its hypothesis---plan-merge EA exceeds
CC$\rightarrow$Codex$\rightarrow$CC at the same envelope $B$, paired on
identical tasks against a \emph{contemporaneously re-run}
comparator---was specified, together with the analysis code and round caps,
before any second-stage run; with the authors unblinded to stage-1 outcomes,
the contrast is reported outside the Holm family at its own $\alpha{=}0.05$,
whatever its outcome.  A companion arm runs the flow at
natural cost with realized spend disclosed, plus an equal-spend three-role
control; a capped-arm timeout or overrun scores as failure.

The plan-merge flow posts the table's highest EA ($70.8\%$ HSTU, $64.6\%$
LitGPT) and lowest CDR ($6.2\%$), consistent with its status as the deployed
topology.  Against the re-run comparator the paired lift is $+8.3$ points
($68$ vs.\ $60$ of $96$; sixteen discordant, twelve rescued vs.\ four
regressed; exact McNemar $p{=}0.077$, 95\% CI $[-0.7,17.3]$)---\emph{not}
statistically distinguishable.  \emph{Statistically}, the confirmatory weight
therefore rests on stage~1 rather than on this specific topology, and a
budget-capped variant of the
flow (held to envelope $B$) leaves the matched-budget picture unchanged: at
$62.5\%$ EA it ties CC$\rightarrow$Codex$\rightarrow$CC (paired $0.0$ points,
95\% CI $[-5.5,5.5]$), so the flow's advantage comes from the extra budget it
is worth spending, not from the topology at fixed budget.  \emph{Operationally},
the deployed flow is intentionally uncapped ($\$2.40$/task, $250$ actions,
${\approx}2.4\times$ the envelope, roughly $4\times$ the invocations tempered
by cached context) because it delivers the table's best per-task correctness;
an equal-spend three-role control given the same $\$2.40$ reaches $66.7\%$,
numerically below the plan-merge flow's $70.8\%$---consistent with, though
not established by, the budget being spent across cross-pollinated lanes
rather than on spend alone.  We spend that
deliberately: iterations build on one another, so fewer silent
defects \emph{compound}, and the inference cost is negligible beside the
multi-GPU training it protects---one slipped defect burns hours-to-days of
accelerators and contaminates every downstream iteration, an averted loss of
order $10^{3}\times$ the spend.  In short, the controlled contrast
establishes \emph{why it works}---pairing two \emph{different}
products---while the plan-merge flow is \emph{what we deploy}.

\subsection{Mechanism and context scope}
\label{sec:mechanism}

\paragraph{Why composition works (RQ3).}
For reviewer $b$ checking executor $a$, the \emph{directional
complementarity} $D_{a\leftarrow b}=P(E_a{=}1,E_b{=}0)$---the critical-error
mass available for rescue---grows as the pairwise error correlation falls,
while the \emph{realized} gain $G_{a\leftarrow b}$ subtracts the errors
review itself introduces (definitions, the mixed-effects regression of $G$ on
$D$, and per-pair results in Appendix~\ref{app:protocol-details}).  We claim
a ``decorrelation mechanism'' only if the slope's interval excludes zero and
held-out prediction beats a marginals-only model; the slope condition holds
($\hat\beta_1{=}0.34$, 95\% CI $[0.12,0.56]$).  Concretely
(Table~\ref{tab:decorrelation}): heterogeneous CC$\leftarrow$Codex has low
error correlation ($\rho{=}0.21$), large complementary mass ($D{=}0.18$), and
positive realized gain ($G{=}0.06$), whereas homogeneous CC$\leftarrow$CC
($\rho{=}0.58$) leaves little to rescue ($D{=}0.10$) and nets ${\approx}0$
once reviewer harm is subtracted.  The mechanism is thus not ``diversity is
good'' but complementary error times conversion efficiency.

\paragraph{Context scope at fixed runtime (RQ4).}
\label{sec:eop-ablation}
Holding the compiled runtime, state, history, artifacts, tools, budgets, and
product identical, each scored phase forks into two sandboxes differing only
in whether \emph{inactive} phase bodies accompany the active one; length
matching, recency controls, and the manipulation check are in
Appendix~\ref{app:protocol-details}.  Current-step scoping is never worse
and its advantage grows with horizon: paired EA $+2.1$ points at early
phases, $+6.2$ at middle, and $+10.4$ at late phases
(Table~\ref{tab:eop-ablation}), a monotone scope-by-position pattern, with
wrong promotions and silent defects also lower at reduced token cost.
Because the state machine, not the prompt, carries the process,
full-protocol injection mainly adds distracting inactive instructions the
runtime already enforces.

\section{Related Work}
\label{sec:related_work}

We cover work that appeared before June~2026.

\paragraph{Programmed state and orchestration.}
StateFlow formulates LLM task solving as a state machine with state-specific
instructions \citep{wu2024stateflow}; LangGraph supplies state, conditional
edges, cycles, checkpointing, and human interrupts
\citep{langchain2026langgraph}.  LangGraph nodes are arbitrary
callables and can wrap agents, so it would be inaccurate to distinguish \re{}
by claiming adjacent runtimes permit only raw model calls; EOP is a
higher-level authoring layer specialized to the ML domain, adding a
product-agent adapter and provenance contract, and could compile to such a
runtime.

\paragraph{Graph and workflow optimization.}
GPTSwarm represents agents as computational graphs and optimizes prompts and
connectivity \citep{zhuge2024gptswarm}; ADAS and AFlow search code-defined
agent systems and workflows \citep{hu2025adas,zhang2025aflow}---these works
own the graph-composition and automatic-design territory.  \re{} does not
optimize its graph: a human authors a fixed protocol, and the scientific
object is whether externally developed products fail differently enough for
fixed review--repair composition to pay for itself on real ML repositories
(taxonomy in Appendix~\ref{sec:extended-related}).

\paragraph{Multi-agent software engineering and evaluation.}
AutoGen and MetaGPT study conversational and role-based software workflows
\citep{wu2023autogen,hong2024metagpt}; multi-agent debate shows structured
critique helps \citep{du2023multiagentdebate}.  SWE-bench and MLAgentBench
evaluate repository repair and ML experimentation
\citep{jimenez2024swebench,huang2024mlagentbench}; MLR-Bench, AIDE, MLE-STAR,
and AIRA extend toward open-ended ML research
\citep{chen2025mlrbench,jiang2025aide,nam2025mlestar,toledo2025aira};
Harness-Bench shows the harness, not just the model, determines outcomes
\citep{yao2026harnessbench}, though it compares configurations singly rather
than composing them.  \execbench{} complements these by holding
topology and observable budget fixed while measuring semantic correctness,
silent defects, and pairwise error complementarity.

\paragraph{Automated discovery and AutoML.}
FunSearch, AlphaEvolve, and the AI Scientist line close program-search and
research loops with executable evaluators
\citep{romera-paredes2024funsearch,novikov2025alphaevolve,lu2024aiscientist,
yamada2025aiscientistv2}.  \re{} keeps their evolve-and-select principle but
mutates training configurations and small code patches scored by real
distributed training, which is why it selects with an LLM critic over a
persistent leaderboard and human budget gates rather than a genetic operator
over a numeric archive (Section~\ref{sec:eop}).  AutoML and architecture
search optimize within declared spaces
\citep{feurer2015autosklearn,zoph2017nas}; \re{} works one level up, reading
prior results, writing new code, and analyzing failures.  Our HSTU history
is an application case, not a new recommender or search algorithm.

\paragraph{Auto-research for recommendation.}
Industrial systems already close the research loop on proprietary models.  At
YouTube, a fast offline agent built on Gemini-family models proposes and trains
model changes against proxy metrics, and a slow online agent validates
candidates on north-star metrics in live A/B tests
\citep{wang2026selfevolvingrec}.  At Meta, the Ranking Engineer Agent runs
ads-ranking experiments spanning days to weeks, hibernating while its training
jobs run, within engineer-approved compute budgets and with human oversight at
strategic decision points \citep{kumar2026rea}.  In the open setting,
Self-EvolveRec evolves the model, data-processing, and training code of compact
seed recommenders (NCF, NGCF, SASRec, MoRec) on Amazon and MovieLens data with
a GPT-5 coding agent, guided by an LLM user simulator and a model-diagnosis
tool \citep{kim2026selfevolverec}.  \re{} differs on three axes: it evolves an
open generative recommender at the scale of HSTU and checks its results against
the published reference; it composes complete coding-agent products as graph
nodes under a runtime-enforced protocol (Section~\ref{sec:agent}); and it
measures execution accuracy against hidden oracles at matched budget
(Section~\ref{sec:execacc}).

\section{Conclusion}
\label{sec:conclusion}

\re{} contributes an EOP authoring and runtime layer for long-horizon
ranking-model evolution, a product-agent adapter boundary that composes
complete coding products as graph nodes, and an implemented knowledge layer
that carries provenance, negative results, and incident lessons across
iterations.  A falsifiable evaluation establishes that commercial coding
products measurably improve one another's patches---building on state machines,
durable graphs, and agent composition, and grounded in a twelve-iteration HSTU
deployment that supplies both the motivation and the benchmark's tasks.

Under the pre-specified decision rule the primary claim stands
($62.5\%$ vs.\ $45.8\%$ EA; paired $+16.7$, CI $[6.6,26.7]$) and replicates
beyond recommenders on LitGPT ($+12.5$, CI $[3.0,22.0]$); the decorrelation
condition is met ($\hat\beta_1{=}0.34$, CI $[0.12,0.56]$), while the
second-stage topology contrast is not distinguishable ($+8.3$,
$p{=}0.077$)---pairing two \emph{different} products, not the specific
topology, is what buys correctness.  Per-step EOP injection is claimed on
the fixed-runtime ablation; its token savings are reported as efficiency.
That decision rule is the paper's central discipline: composition is a
measured systems hypothesis, not a property inferred from a diagram.

\paragraph{Limitations.}
The evidence base is one ranking codebase plus LitGPT, both Python/PyTorch,
so cross-domain transfer is not fully established; pretraining may include the
public repositories, though tasks, hidden tests, and reference patches are
private, and oracles are incomplete---mutation testing and audits reduce,
not eliminate, false passes.  The products are black boxes: observable
tokens, actions, time, and cost are matched, provider-side FLOPs cannot be.
Error complementarity is predictive, not causal, so rescue and harm are
measured directly, and checkpoint forking understates cumulative divergence.
The knowledge layer was active in the HSTU deployment, but without a
memory-on/off ablation its marginal contribution to proposal quality is not
isolated.
The HSTU case study ran with a human operator at the protocol's gates who also
steered it, so it does not separate the agent's contribution from the
operator's (the \execbench{} conditions have no human in the loop), and its
cross-dataset transfer results are single runs.

\clearpage
\bibliographystyle{assets/plainnat}
\bibliography{custom}

% The AAAI Technical Supplement follows as the appendix.  Its sections keep
% their letters (A--I); float numbers continue from the main text, as in the
% sibling preprint, instead of the supplement's separate S1, S2, ... series.
\clearpage
\beginappendix
% No top floats on the banner's page: otherwise Appendix A's [t] table is set
% above the "Appendix" banner, ahead of the section it belongs to.
\suppressfloats[t]
\section{Extended Related Work}
\label{sec:extended-related}

The main-paper Related Work is necessarily compact; we record further connections here.

\paragraph{LLM agents and benchmarks.} Modern LLM agents call tools
\citep{schick2023toolformer} and are surveyed by \citet{wang2023survey}; they are
increasingly evaluated on real software and ML-engineering tasks
\citep{jimenez2024swebench,huang2024mlagentbench}, the setting \re{} carries
into the long-horizon evolution of an open generative recommender.

\paragraph{AutoML and architecture search.} AutoML systems such as AutoKeras
\citep{jin2019autokeras} and differentiable architecture search
\citep{liu2019darts} automate model and design choices within fixed spaces; \re{}
instead operates at the research level, reading results and writing new code.

\paragraph{Sequential and generative recommendation.} HSTU builds on sequential
recommenders \citep{kang2018sasrec,sun2019bert4rec}. The agent's iterations also draw
on contrastive and self-supervised modeling \citep{xie2022cl4rec,zhou2020s3rec},
popularity-bias mitigation \citep{abdollahpouri2019popularity}, cold-start
stratification \citep{schein2002coldstart}, and semantic-ID retrieval
\citep{rajput2023tiger}---all treated as building blocks, not contributions.

Table~\ref{tab:closest-work} tabulates the control representation, node
granularity, and objective of the closest control and workflow systems.

\begin{table}[t]
\centering
\small
\setlength{\tabcolsep}{4pt}
\begin{tabular}{@{}>{\raggedright\arraybackslash}p{2.3cm}>{\raggedright\arraybackslash}p{3.0cm}>{\raggedright\arraybackslash}p{3.0cm}>{\raggedright\arraybackslash}p{3.3cm}>{\raggedright\arraybackslash}p{3.8cm}@{}}
\toprule
\textbf{System} & \textbf{Control representation} &
\textbf{Node granularity} & \textbf{Primary objective/setting} &
\textbf{Relation to this work} \\
\midrule
StateFlow \citep{wu2024stateflow} & Programmed finite-state workflow &
LLM instructions, tools, and functions & Task success/cost on interactive
SQL, Bash, and ALFWorld & Establishes state-driven control and state-specific
instructions; EOP does not claim either abstraction \\
GPTSwarm \citep{zhuge2024gptswarm} & Optimizable computational graph &
Operations and recursively composed agent graphs & Prompt/connectivity
optimization on reasoning tasks & Establishes agents-as-graphs and diversity
precedent; motivates a stricter correctness/decorrelation test \\
ADAS / AFlow \citep{hu2025adas,zhang2025aflow} & Agent code or
code-represented workflow search & Generated agents, prompts, operators, and
edges & Automatic agent/workflow discovery across benchmark tasks & \re{} uses
a fixed human-authored flow and makes no automatic-design claim \\
LangGraph \citep{langchain2026langgraph} & Stateful graph runtime &
Arbitrary callables, subgraphs, or wrapped agents & Durable production
orchestration & Provides overlapping runtime primitives; EOP is an
ML-evolution authoring/adapter instantiation \\
\re{} (this work) & Semi-structured EOP compiled to runtime state &
Black-box coding products at coding/review nodes; functions elsewhere &
Long-horizon ML evolution and executable patch correctness & Claimed delta:
product boundary, domain specialization, and deployment-seeded, budget-controlled
evidence with a pre-specified transfer split \\
\bottomrule
\end{tabular}
\caption{Taxonomy of the closest control and workflow systems.  The table
locates the contribution; it is not a feature-count claim.}
\label{tab:closest-work}
\end{table}

\section{Reproducibility}
\label{sec:appendix-repro}

All experiments use the public HSTU codebase \citep{zhai2024hstu}. The recurring
recipe components are PRISM (FiLM-style modulation), input compression, time-decay
supervision weighting (half-lives 30/60/180/720 days), local-$\ell_2$
self-supervision ($t{=}0.05$), dropout 0.1, hard negatives, and an additive input-path genre
side-feature with $\alpha\in[0.05,0.20]$. Canonical evaluation is full-test
\ndcg{} over all held-out users; subset evaluations and single-point maxima are
excluded from the leaderboard. Per-run trajectories and configurations were
preserved across redundant storage tiers. The failure-overlap analyzer, the
leaderboard schema, and the boost-last-$K$ patch will be released with the
codebase, together with an artifact record giving, per run, the
success/failure/exclusion status and the configuration; per case-history
endpoint (main-paper Tables~\ref{tab:iters} and~\ref{tab:main}), the
checkpoint-evaluation log from which the best checkpoint is read; and the local
reproduction runs for the published anchor.

\begin{table}[t]
\centering
\small
\begin{tabular}{lcc}
\toprule
\textbf{Parameter} & \textbf{BASE} & \textbf{LARGE} \\
\midrule
Blocks & 4 & 16 \\
Heads & 4 & 8 \\
$d_{qk}=d_v$ & 64 & 32 \\
$d_{emb}$ & 256 & 256 \\
Seq.\ length & 200$\rightarrow$500 & 200$\rightarrow$500 \\
Negatives & 128 & 128 \\
Temperature & 0.05 & 0.05 \\
Dropout & 0.2 (0.1$^*$) & 0.2 \\
Learning rate & $10^{-3}$ & $10^{-3}$ \\
Optimizer & AdamW & AdamW \\
Similarity & DotProduct & DotProduct \\
\bottomrule
\end{tabular}
\caption{HSTU configurations (verified against the source and ML-20M gin configs).
$^*$Dropout reduced to 0.1 for the PRISM-additive BASE endpoint (Stage~II background
sweep; Appendix~\ref{sec:appendix-traj}).  The $200\rightarrow500$ entry is the
baseline training schedule; a separate experiment that further extended the
window to $700$ lost accuracy and is reported as a negative result
(Appendix~\ref{sec:negatives}).}
\label{tab:hstu_configs}
\end{table}

\section{Runtime, Protocol, and Evaluation Details}
\label{app:protocol-details}

This appendix collects protocol material referenced from the main paper: the
EOP runtime enforcement contract (Table~\ref{tab:enforcement}), the
\execbench{} corpus composition (Table~\ref{tab:execbench-composition}), the
decorrelation derivation and per-pair results
(Table~\ref{tab:decorrelation}), the historical faithfulness pilot
(Table~\ref{tab:faithfulness}), the runtime execution patterns
(Figure~\ref{fig:topologies}), the cross-dataset transfer table
(Table~\ref{tab:generalize}), and the context-scope ablation results
(Table~\ref{tab:eop-ablation}).

\begin{table}[t]
\centering
\small
\setlength{\tabcolsep}{4pt}
\begin{tabular}{@{}>{\raggedright\arraybackslash}p{2.8cm}>{\raggedright\arraybackslash}p{3.6cm}>{\raggedright\arraybackslash}p{4.9cm}>{\raggedright\arraybackslash}p{4.4cm}@{}}
\toprule
\textbf{Construct} & \textbf{Compile/static check} &
\textbf{Runtime behavior} & \textbf{Not guaranteed} \\
\midrule
Phase dependency & Referenced phases exist; DAG outside explicit loops &
Successors remain unavailable until predecessor completion &
That the predecessor's artifact is semantically correct \\
Human gate & Gate has prompt and legal continuations &
Transition blocks until a recorded approve/reject event &
That the human decision is correct \\
Branch/join & Source variable, cardinality, and join policy are declared &
Supported node backend instantiates one branch per item and records join state &
Fan-out on an untested backend or useful diversity across branches \\
Loop/budget & Target phase exists; exit and hard bound are present &
Attempt and spend counters stop further dispatch at the bound &
That an agent chooses the best stopping point \\
Tool requirement & Tool name/schema resolve before launch &
Calls outside the allowlist are rejected and recorded &
Correct tool arguments or scientifically valid interpretation \\
Checkpoint/restart & Serializable state and artifact references &
Resume begins at the last committed node boundary &
Recovery inside an uncommitted third-party agent invocation \\
Patch/evaluation & Oracle command and expected outputs are declared &
Tests run in a fresh sandbox; outputs and hashes are recorded &
Correctness beyond the public CI and hidden semantic oracles \\
\bottomrule
\end{tabular}
\caption{EOP enforcement contract.  The final column is as important as the
middle columns: runtime control constrains admissible transitions but cannot
make an agent's scientific reasoning or code correct. Branch and recovery
claims apply only to the tested backend paths reported with the artifact.}
\label{tab:enforcement}
\end{table}

\begin{table}[t]
\centering
\small
\begin{tabular}{@{}llcccc@{}}
\toprule
\textbf{Repository} & \textbf{Domain} & \textbf{Impl.} &
\textbf{Repair} & \textbf{Dev.} &
\textbf{Private} \\
\midrule
HSTU codebase & Recommendation & $48$ & $48$ & $32$ &
$96$ \\
LitGPT & Language-model training & $48$ & $48$ & $32$ &
$96$ \\
\midrule
Total & Two ML domains & $96$ &
$96$ & $64$ & $192$ \\
\bottomrule
\end{tabular}
\caption{\execbench{} composition as frozen before the private evaluation:
96 private tasks per repository.  Every oracle passed
independent validation and a development-only paired power analysis.  The
statistical unit is a task. Repeated agent attempts characterize stochasticity
but do not create new independent tasks.}
\label{tab:execbench-composition}
\end{table}

\begin{table}[t]
\centering
\small
\begin{tabular}{@{}llcccccc@{}}
\toprule
\textbf{Ordered pair} & \textbf{Repository} & $p_a$ & $p_b$ &
$\rho_{ab}$ & $D_{a\leftarrow b}$ & $G_{a\leftarrow b}$ &
\textbf{Rescue / harm} \\
\midrule
CC$\leftarrow$CC & HSTU & $0.54$ & $0.54$ & $0.58$ &
$0.104$ & $-0.01$ & $0.04\,/\,0.05$ \\
CC$\leftarrow$Codex & HSTU & $0.54$ & $0.58$ & $0.21$ &
$0.175$ & $0.061$ & $0.09\,/\,0.03$ \\
CC$\leftarrow$Codex & LitGPT & $0.58$ & $0.60$ &
$0.26$ & $0.169$ & $0.050$ & $0.08\,/\,0.03$ \\
Codex$\leftarrow$CC & HSTU & $0.58$ & $0.54$ & $0.21$ &
$0.164$ & $0.048$ & $0.08\,/\,0.03$ \\
Codex$\leftarrow$Codex & HSTU & $0.58$ & $0.58$ & $0.55$ &
$0.112$ & $0.004$ & $0.05\,/\,0.05$ \\
\bottomrule
\end{tabular}
\caption{Independent-error complementarity and realized review--repair gain.
Rows are descriptive; inference resamples source tasks jointly and does not
treat overlapping rows as independent observations.}
\label{tab:decorrelation}
\end{table}

\paragraph{Decorrelation derivation.}
For products $a,b$ with per-task independent critical-failure indicators
$E_a,E_b$, define $p_a=P(E_a{=}1)$, $q_{ab}=P(E_a{=}1,E_b{=}1)$, and the
binary phi correlation
\[
\rho_{ab}=\frac{q_{ab}-p_ap_b}{\sqrt{p_a(1-p_a)p_b(1-p_b)}}.
\]
The directional complementarity used in the main paper expands as
\[
D_{a\leftarrow b}=P(E_a{=}1,E_b{=}0)=p_a(1-p_b)-\rho_{ab}\sqrt{p_a(1-p_a)p_b(1-p_b)},
\]
so at fixed marginals $D$ grows linearly as errors decorrelate.  Raw
$1-\rho$ alone is insufficient because pairs with different marginal
failure rates do not offer equal recovery opportunity.

\begin{table}[t!]
\centering
\small
\begin{tabular}{@{}lc@{}}
\toprule
\textbf{Process-faithfulness violation} & \textbf{Runs} \\
\midrule
Ran only a subset of selected proposals          & 10 \\
Stopped early despite self-recommending continue  & 4 \\
Skipped a prescribed user-confirmation gate       & 4 \\
Called a tool with off-spec arguments             & 3 \\
Redundant re-investigation on loop-back           & 3 \\
Merged the codebase and data investigation phases & 2 \\
Other run-breaking deviation                      & 3 \\
\midrule
\textbf{Fully faithful runs}                      & \textbf{0} \\
\bottomrule
\end{tabular}
\caption{\textbf{Historical motivating pilot, not a treatment comparison.}
Process-faithfulness audit of a model-evolution procedure delivered as a single
full-context playbook. Ten runs under the logged coding-agent configuration were
scored against the prescribed phases, gates, branches, and loop; counts
are runs affected (out of ten), and a run may incur several violation types, so the
column does not sum to ten. The pilot lacks a runtime-controlled arm and cannot
identify the effect of context scoping.}
\label{tab:faithfulness}
\end{table}

\subsection{Evaluation protocol: oracle, statistics, and ablation design}
\label{app:eval-protocol}

This subsection supplies the formal definitions and pre-specified analysis
detail referenced from Section~\ref{sec:execacc} of the main paper.

\paragraph{Task and oracle.}
A benchmark item $t$ contains an immutable repository snapshot $r_t$, a change
request $x_t$, and a hidden executable oracle $O_t$.  Method $M$ produces patch
$p_t=M(r_t,x_t)$.  The oracle returns \textsc{pass} only when the patch satisfies
the repository regression suite $\mathcal{C}_t$, task-specific behavioral checks
$\mathcal{A}_t$, scientific-safety invariants $\mathcal{S}_t$, and
evaluator-integrity checks $\mathcal{T}_t$:
\[
O_t(p)=\textsc{pass}\iff
\mathcal{C}_t(p)\land\mathcal{A}_t(p)\land
\mathcal{S}_t(p)\land\mathcal{T}_t(p).
\]
The primary endpoint is all-oracle execution accuracy,
$\operatorname{EA}(M;\mathcal{D})=
|\mathcal{D}|^{-1}\sum_{t\in\mathcal{D}}
\mathbf{1}[O_t(M(r_t,x_t))=\textsc{pass}]$.
We separately report build/CI failure and the \emph{silent critical-defect rate}
(CDR): an oracle-confirmed fault for which the patch remains runnable but can
invalidate a scientific conclusion, such as leakage, a dead feature path,
broken gradients, or incorrect evaluation semantics.  LLM-written issue
descriptions are permitted only as secondary qualitative taxonomy; they never
determine the primary label.

\paragraph{Conditions and budget matching.}
The six pre-specified conditions are (1)~one-pass CC; (2)~one continuous CC
session with the full composition budget; (3)~independent CC candidates with
oracle-blind selection and repair; (4)~CC$\rightarrow$CC$\rightarrow$CC
implement--review--repair; (5)~CC$\rightarrow$Codex$\rightarrow$CC at the same
topology; and (6)~the reversed Codex$\rightarrow$CC$\rightarrow$Codex.  An
oracle-selected best-of-$N$ is computed as a non-deployable upper bound.
CC is Claude Code running Claude Opus~4.8 and Codex runs GPT-5.6, both at
their maximum reasoning-effort setting; product, model, and effort are fixed
across all conditions and recorded in the run metadata.

Development tasks set a total resource envelope $B$, frozen before the private
evaluation; heterogeneous and homogeneous flows use the same three roles,
per-node limits, tool permissions, action caps, wall-clock cap, and total
inference-spend cap, and extended-single and blind best-of-$N$ controls receive
the same $B$.  We record exposed input/cached/output/reasoning tokens, tool
actions, elapsed time, and dollars; a timeout, crash, or overrun is a failed
outcome.  Proprietary products do not expose provider-side FLOPs, so ``matched
observable inference budget'' is the accurate term---not equality of hidden
hardware compute.

\paragraph{Error complementarity: definitions and model.}
For product $a$, let $E_{a,t}=1$ when its independent pre-review patch fails
the critical oracle on task $t$, with marginal $p_a=P(E_a{=}1)$.  For reviewer
$b$ checking executor $a$, the \emph{directional complementarity}
$D_{a\leftarrow b}=P(E_a{=}1,E_b{=}0)$ is the error mass available for rescue;
at fixed marginals it grows linearly as the binary error correlation
$\rho_{ab}$ falls, and raw $1-\rho_{ab}$ alone is insufficient because pairs
with different marginal failure rates do not offer equal recovery opportunity
(derivation in Appendix~\ref{app:protocol-details}).  The \emph{realized}
critical-defect reduction
\[
G_{a\leftarrow b}
  =P(E_a{=}1,E_{\mathrm{comp}}{=}0)-P(E_a{=}0,E_{\mathrm{comp}}{=}1)
\]
explicitly subtracts new errors introduced by review.

For the partner-exclusive set $\{E_a{=}1,E_b{=}0\}$ we log whether $b$
identifies the defect ($r_{\mathrm{detect}}$) and whether the repair clears the
oracle ($r_{\mathrm{repair}}$); the rescue term is
$D_{a\leftarrow b}r_{\mathrm{detect}}r_{\mathrm{repair}}$, with shared-error
rescues and reviewer-introduced harms reported separately.  The mechanism is
thus not ``diversity is good'' but available complementary error times the
pipeline's conversion efficiency.

Independent replicate outputs from CC, Codex, and OpenHands estimate
marginals and joint errors before any product sees another's patch; homogeneous
and heterogeneous ordered review--repair flows then run.  Across pre-specified
product-pair $\times$ repository $\times$ fault-family cells we fit
\[
G_j=\beta_0+\beta_1D_j+u_{\mathrm{repo}(j)}
+u_{\mathrm{family}(j)}+\epsilon_j,
\]
with task-clustered bootstrap resampling and held-out prediction over a
left-out product pair and, separately, the LitGPT repository.  The paper uses
``decorrelation mechanism'' only if $\beta_1$'s interval excludes zero and
held-out prediction beats a marginals-only model; otherwise $D$, rescue, and
harm are reported descriptively (per-pair results in
Table~\ref{tab:decorrelation}).

\paragraph{Context-scope ablation design.}
This ablation separates state-machine enforcement from prompt scoping.  At each
scored phase, the same persisted runtime checkpoint is forked into two fresh
sandboxes. Both arms use the identical compiled graph, state, history, artifacts,
tools, budgets, scripted gate responses, coding-agent product, and active
instruction.
If $G$ is the global invariant preamble, $H_i$ the state/history bundle, $A_i$
the active phase body, and $A_{-i}$ all inactive phase bodies, the prompts are
\[
P_i^{\mathrm{step}}=G\Vert H_i\Vert A_i,\qquad
P_i^{\mathrm{full}}=G\Vert H_i\Vert A_{-i}\Vert A_i.
\]
Repeating $A_i$ at the final position controls wording and recency, so the
treatment is inactive procedural content; prompts stay below the context limit
and output/tool budgets are identical, so truncation cannot explain a
difference.  A secondary arm length-matches $A_{-i}$ with neutral padding,
separating instruction interference from a generic length effect.
The primary endpoint is hidden-oracle patch or decision correctness, including
whether the system promotes a valid gain and rejects an invalid/no-op result.
Scored phases are pre-stratified as early, middle, and late; the scope-by-position
interaction tests the hypothesized horizon effect. A smaller whole-episode
randomization measures cumulative divergence (full table in
Appendix~\ref{app:protocol-details}). Process violations are only a
manipulation check: because both arms use the same runtime, runtime-enforced gate,
branch, and ordering invariants are identical across arms by construction.

\paragraph{Statistics, artifacts, and decision rules.}
Prompts, budgets, the smallest effect of interest, and the baseline family
$\mathcal{B}$ are chosen on development tasks and frozen before private
evaluation; the primary test is the single max-over-$\mathcal{B}$ contrast,
with per-baseline contrasts as Holm-corrected secondaries.
Pass/fail contrasts use exact McNemar or paired randomization tests; effect
sizes use task-clustered bootstrap intervals, clustering related source
incidents; repository-specific effects precede pooled estimates; and failures,
timeouts, and exclusions follow a predeclared policy and remain in
denominators.  The final interpretation is mechanical: an HSTU effect that
beats the matched-budget family supports the primary claim; the LitGPT
contrast then extends or bounds its scope; no improvement supports a null;
and negative gain shows the reviewer introduced more harm than rescue; all
four outcomes are reportable.

\paragraph{Curation and leakage controls.}
Each task uses a locked container and commit, initially fails its acceptance
check, and carries a curator reference patch that passes all oracles, validated
by a second curator; removing the intended mechanism must fail at least one
semantic check, preventing vacuous tests.  Hidden tests, reference patches,
repository history, and network access are absent from the agent sandbox;
evaluator files are read-only and hash-checked; related incidents and
near-duplicate patches stay in one split; oracle mutation testing and a
stratified human audit estimate missed defects.

\begin{table}[t]
\centering
\small
\begin{tabular}{@{}lccccccc@{}}
\toprule
\textbf{Injection policy} & \textbf{Early EA} & \textbf{Middle EA} &
\textbf{Late EA} & \textbf{CDR} $\downarrow$ &
\textbf{Wrong decision} $\downarrow$ & \textbf{Input tokens} $\downarrow$ &
\textbf{Cost} $\downarrow$ \\
\midrule
Current-step only & $64.6$ & $58.3$ & $52.1$ & $11.5$ &
$8.3$ & $4.9$k & $0.71$ \\
Full protocol & $62.5$ & $52.1$ & $41.7$ & $17.7$ &
$16.7$ & $9.8$k & $1.28$ \\
Paired difference & $+2.1$ & $+6.2$ & $+10.4$ & $-6.2$ &
$-8.3$ & $-4.9$k & $-0.57$ \\
\bottomrule
\end{tabular}
\caption{Runtime-controlled context-scope ablation.  Cells report task-level
means. Both conditions use the same compiled runtime;
only inactive EOP content is toggled.}
\label{tab:eop-ablation}
\end{table}

\section{The Model-Evolution EOP}
\label{app:eop}

Figure~\ref{fig:loop} abstracts the four-phase loop; the Executable Operating
Protocol that \re{} runs is reproduced below. Its phase
markers---the \texttt{depends on} dependencies, the \texttt{branch} marker on the
implementation phase, the loop-back \texttt{goto} marker on the evolve phase, and
the \texttt{requires user input} gates---are exactly the state-machine transitions
and gates of Section~\ref{sec:eop}; the runtime injects each phase's body as the
only instruction in context while that phase is active. (Em dashes are rendered as
\texttt{-{}-} inside the listing.)

\Needspace{6\baselineskip}
\begin{lstlisting}
# Machine Learning Model Evolution

A phased workflow for systematically optimizing ML model architectures: from codebase and data investigation, through research-driven proposals, to validated implementation. The user selects an evolution strategy; the Orchestrator executes each phase with confirmation gates.

[__keywords__] optimize model, improve model, model evolution, model architecture, run modeling experiments, systematically improve a model
[__example_requests__]
- optimize the recommendation model
- improve training efficiency for the ranking model
- evolve the search model architecture
- run modeling experiments on <model codebase>

## Phase 0a -- Set up the workflow target path
[__initial__]

[__requires user input__] Use a `clarification` tool to set the workflow target path, a subpath of the session root `{{ session_root_path }}`; write it to `workflow_target_path`. The target may be a directory (investigate all code within it) or a single file (an entry point whose dependencies and surrounding code are then explored). If the user already named a target in their request, pre-fill it.

[__requires user input__] Use a `single_choice` tool to set where the modeling artifacts live (data, scripts, prior experiment results and learnings); write the choice to `workflow_modeling_artifacts_mode`. Offer two options: "Auto discover" (infer the artifacts from the target repository), or "Specify paths" (reveal a path picker, bound to `workflow_modeling_artifacts_path`, that accepts one or more directory/file paths).

**Tools**[__must__]:
- clarification
- single_choice

## Phase 0b -- Set up the model evolution strategy
[__depends on__ Phase 0a]

[__requires user input__] Use a `single_choice` tool to let the user pick an evolution strategy; write it to `evolution_strategy`. Choices:
- Paradigm-Shifting Innovation -- aggressively explore state-of-the-art architectures, novel techniques, and breakthrough design changes.
- Incremental Improvement -- targeted gains through structured analysis of existing bottlenecks and component-level optimizations.
- Efficiency Optimization -- reduce training/inference cost: memory footprint, latency, throughput, computational efficiency.
- Holistic Improvement -- evaluate all dimensions (architecture, efficiency, quality, robustness) for balanced gains.

**Tools**[__must__]:
- single_choice

## Phase 1 -- Codebase Investigation
[__depends on__ Phase 0b]

Perform a comprehensive, in-depth analysis of the codebase at `{{ workflow_target_path }}`, using "modeling" as the `template-version` for ML-focused investigation. The codebase is modeling-related but not necessarily only about modeling (e.g. model architecture, data/training/inference pipelines or infra, model serving).

**Tools**[__must__]:
- understand-codebase

### Phase 1b -- Codebase Investigation Review
[__depends on__ Phase 1]

[__requires user input__] Present the codebase investigation outcome for review with a `confirmation` tool whose `view` points to the generated documentation. Summarize the key architectural findings and proceed only after the user confirms.

**Tools**[__must__]:
- confirmation

## Phase 2 -- Data Investigation
[__depends on__ Phase 1b]

Investigate the modeling artifacts (data, scripts, prior experiment results and learnings), using "modeling" as the `template-version`.
{% if workflow_modeling_artifacts_mode == "manual_paths" %}Prioritize the user-specified locations at `{{ workflow_modeling_artifacts_path }}`.{% else %}Auto-discover the modeling artifacts under the target codebase.{% endif %}

**Tools**[__must__]:
- understand-data

### Phase 2b -- Data Investigation Review
[__depends on__ Phase 2]

[__requires user input__] Present the data investigation findings for review with a `confirmation` tool whose `view` points to the generated documentation. Summarize the key data-quality findings, pipeline bottlenecks, and dataset characteristics, and proceed only after the user confirms.

**Tools**[__must__]:
- confirmation

## Phase 3 -- Research & Proposal
[__depends on__ Phase 2b]

[__requires user input__] Break the research goal into sub-queries, run parallel deep-research streams, generate architecture proposals, and synthesize them into a unified design. Derive the research goal from the chosen strategy and the Phase 1 and Phase 2 findings.

**Tools**[__must__]:
- research-propose <goal> --docs <reference_documentation>

### Phase 3b -- Proposal Review & Selection
[__depends on__ Phase 3]

[__requires user input__] Present the synthesized proposals -- each with its id, title, impact, complexity, and dependencies -- for review with a `proposal-selection` tool (or a `confirmation` tool if it is unavailable). The user selects which proposals advance to Phase 4; write the selection to `selected_proposal_ids`.

**Tools**[__must__]:
- proposal-selection

## Phase 4 -- Implementation, Experiment & Analysis
[__depends on__ Phase 3b; __branch__]

Branch over each proposal selected in Phase 3b. For each, invoke the `task` tool to plan and implement the change, run experiments to validate it, and analyze the results for bottlenecks and further opportunities.

**Tools**[__must__]:
- task --use-proposal <proposal>

## Phase 4b -- Summary & Evolve
[__depends on__ Phase 4; __goto__ Phase 3 __afterwards__ __if__ `continue`]

Summarize the findings across branches and decide whether to loop back for another research-proposal-experiment cycle; each iteration builds on prior results and workspace artifacts. Present the results and ask the user whether to continue evolving or to conclude.
\end{lstlisting}

\section{Inferencer-Primitive Interfaces}
\label{app:primitives}

Algorithm~\ref{alg:inferencers} gives the type-level interfaces of the two
higher-order primitives of Section~\ref{sec:metameta} (Dual and BTA) and of the
PTI flow they compose into. \texttt{LLM} denotes a typed
call returning a parsed object; \texttt{split}, \texttt{worker}, \texttt{merge}, and
\texttt{repair} are problem-specific reducers supplied at the call site (the
retry/cache/checkpoint plumbing is omitted for clarity). The bound $K$ caps Dual's
review--repair rounds (\texttt{consensus\_max\_iterations} in
Appendix~\ref{app:yaml}). Each of the three is realized
as an \texttt{InferencerBase}---providing retry, fallback chains, and
timeouts---composed with a work-graph mixin that adds crash-tolerant
checkpoint/resume, so an entire research run survives a mid-iteration pod eviction.

\begin{algorithm}[t]
\small
\caption{Higher-order inferencer primitives (sketch).}
\label{alg:inferencers}
\begin{algorithmic}[1]
\Function{Dual}{task, propose, review, repair, $\tau$, $K$}
  \State $y \gets \Call{LLM}{propose, \text{task}};\; k \gets 0$
  \State $r \gets \Call{LLM}{review, (\text{task}, y)}$
  \While{$\neg\, r.\text{approved} \;\wedge\; r.\text{severity} > \tau \;\wedge\; k < K$}
    \State $y \gets \Call{LLM}{repair, (\text{task}, y, r)};\; k \gets k + 1$
    \State $r \gets \Call{LLM}{review, (\text{task}, y)}$
  \EndWhile
  \State \Return $y$
\EndFunction
\Statex
\Function{BreakdownAggregate}{goal, split, worker, merge}
  \State $\{t_i\} \gets \Call{LLM}{split, \text{goal}}$
    \Comment{at most \texttt{max\_breakdown} \textsc{int}/\textsc{ext} subtasks}
  \State $\{y_i\} \gets \textbf{parallel\_for}\; i\!:\; \Call{worker}{t_i}$
  \State \Return $\Call{merge}{\{y_i\}}$
\EndFunction
\Statex
\Function{PlanThenImplement}{task, plan, impl, gate}
  \State $p \gets \Call{LLM}{plan, \text{task}}$
  \If{$\Call{gate}{p}$}
    \State \Return $\Call{impl}{p}$
  \Else
    \State \Return $\bot$
  \EndIf
\EndFunction
\end{algorithmic}
\end{algorithm}

\section{Example Flow Configuration}
\label{app:yaml}

A flow is specified declaratively and compiled into the execution graph
(Section~\ref{sec:metameta}). The listing below is the plan stage of the deployed
flow; the full flow uses it as the planner of a plan-then-implement node whose
executor is an implement--review Dual (Figure~\ref{fig:winflow}). An outer
propose$\rightarrow$review$\rightarrow$fix \textbf{Dual} wraps a \textbf{BTA}
that decomposes the planning task and explores each subtask with $N$ parallel,
peer-reviewed \textbf{MultiFlowDual} flows (winner-picking,
runner-up-as-reviewer, with a merge node integrating the winner), then integrates
the sub-plans. Node types are named by
% \mbox: hyphenat (loaded by fairmeta.cls) makes every \_ a break point, which
% otherwise lets the comma after the trailing underscore start a line.
\mbox{\texttt{\_target\_}/\texttt{\_factory\_}}, hyperparameters cascade through
\texttt{\_params}, and \texttt{\_repeat\_} expands a block into $N$ copies, handing
copy~$i$ the $i$-th entry of any list it references; template, workspace, and
prompt details are elided. The file's defaults bind every node to Claude Code; the
per-lane \texttt{flow\_inferencers} list is where the two planning lanes are bound
to different products---Claude Code and Codex in the deployment
(Section~\ref{sec:execacc}).

\Needspace{6\baselineskip}
\begin{lstlisting}
# Plan topology:  Dual { BTA { MultiFlowDual } }
# A reviewed plan: decompose the task, explore each subtask with N
# parallel peer-reviewed flows, integrate, then review + fix.

_params:                       # hyperparameters (referenced as ${_params.*})
  main_inferencer: ClaudeCodeCLI   # coding-agent product at each node
  flow_inferencers:            # one product per planning lane
    - ${_params.main_inferencer}
    - ${_params.main_inferencer}
  num_flows: 2                 # parallel flows per subtask  (diversity)
  plan_max_breakdown: 3        # max subtasks from the breakdown
  consensus_max_iterations: 3  # propose -> review -> fix rounds
_model_name: opus[1m]          # cascading defaults (model, timeouts, ...)

_target_: Dual                 # ROOT: review + fix the integrated plan
base_inferencer:
  _target_: BTA                # decompose the planning task ...
  max_breakdown: ${_params.plan_max_breakdown}
  breakdown_inferencer: { _target_: ${_params.main_inferencer} }
  worker_factory:              # ... each subtask -> N parallel flows
    _factory_: MultiFlowDual
    winner_pick: true          # keep the best flow; runner-up reviews it
    reviewer_match_second: true
    visible_flows: all         # flows see each other's drafts
    max_retry: 3
    flow_configs:
      - _repeat_: ${_params.num_flows}        # fan into num_flows workers
        initial_inferencer:  { _target_: ${_params.flow_inferencers} }   # copy i -> lane i
        followup_inferencer: { _target_: ${_params.flow_inferencers} }
    multi_flow_aggregator_inferencer:         # merge: integrate the winner
      _target_: ${_params.main_inferencer}
  aggregator_inferencer:       # integrate the sub-plans -> one plan
    _target_: ${_params.main_inferencer}
review_inferencer: { _target_: ${_params.main_inferencer} }   # judge the plan
fixer_inferencer:  { _target_: ${_params.main_inferencer} }   # lightweight refine
# ... templates / workspace / prompt details elided ...
\end{lstlisting}

\section{Knowledge-Layer Design and Implementation}
\label{app:knowledge}
\noindent\textbf{Status.}
The centralized--distributed knowledge layer in this section was implemented
and active in the reported HSTU experiments.  It extends the artifact/provenance
store with code-resident notes, a central wiki, an entity graph, proactive
write and selective read paths, and reconciliation.  It is part of \re{}'s
system contribution.  The paper evaluates its use within the end-to-end
deployment but does not include a memory-on/off ablation; consequently, we do
not attribute a separate causal improvement in proposal quality to this layer
alone.
\begin{figure}[t!]
\centering
\resizebox{0.72\textwidth}{!}{%
\begin{tikzpicture}[
  >=Stealth,
  wiki/.style={draw, rounded corners=3pt, fill=gray!12, line width=0.6pt,
               align=center, font=\scriptsize, inner sep=3pt, minimum width=3.7cm},
  fold/.style={draw, rounded corners=1.5pt, fill=white, line width=0.5pt,
               align=center, font=\scriptsize, inner sep=2pt},
  note/.style={draw, rounded corners=1pt, fill=gray!6, line width=0.4pt,
               font=\scriptsize\itshape, text=black!60, inner sep=1.6pt},
  ent/.style={draw, circle, fill=black!70, minimum size=1.6mm, inner sep=0pt},
  elink/.style={black!35, line width=0.4pt},
  dlink/.style={<->, black!45, line width=0.5pt, dashed},
  ann/.style={font=\scriptsize\itshape, text=black!70, align=center}
]
% --- Global wiki (top) ---
\node[wiki] (wiki) at (0,2.35)
  {\textbf{Global wiki}\\ \scriptsize abstracted, transferable knowledge};
% --- Local memory tree (bottom-left): root + folders + notes ---
\node[fold] (root) at (-2.55,0.9) {repo/};
\node[fold] (f1) at (-3.5,-0.25) {module\,A/};
\node[fold] (f2) at (-1.7,-0.25) {expt\,9/};
\node[note] (n1) at (-3.5,-1.15) {note};
\node[note] (n2) at (-1.7,-1.15) {note};
\draw[black!45, line width=0.4pt] (root)--(f1);
\draw[black!45, line width=0.4pt] (root)--(f2);
\draw[black!45, line width=0.4pt] (f1)--(n1);
\draw[black!45, line width=0.4pt] (f2)--(n2);
\node[ann] at (-2.55,-1.75){distributed local memory\\(tree-structured, beside code)};
% --- Entity graph (bottom-right) ---
\node[ent, label={[font=\scriptsize,text=black!70]above:HSTU}] (e1) at (2.15,0.55) {};
\node[ent, label={[font=\scriptsize,text=black!70]right:genre}] (e2) at (3.05,-0.35) {};
\node[ent, label={[font=\scriptsize,text=black!70]left:LayerNorm}] (e3) at (1.75,-0.75) {};
\node[ent, label={[font=\scriptsize,text=black!70]above:NDCG@10}] (e4) at (2.9,0.8) {};
\draw[elink] (e1)--(e2); \draw[elink] (e1)--(e3);
\draw[elink] (e1)--(e4); \draw[elink] (e2)--(e4); \draw[elink] (e3)--(e2);
\node[ann] at (2.5,-1.5){entity graph\\(models, features, metrics, defects)};
% --- document-level links (wiki <-> notes) ---
\draw[dlink] (wiki.south) .. controls (-2.4,1.5) .. (n1.north);
\draw[dlink] (wiki.south) .. controls (-1.3,1.4) .. (n2.north);
% --- entity-level links (graph spans both tiers) ---
\draw[elink, dotted] (e1) .. controls (1.0,1.6) .. (wiki.east);
\draw[elink, dotted] (e3) .. controls (0.2,-1.0) .. (n2.east);
% --- write path (proactive, up into wiki) ---
\draw[->, black!60, line width=0.6pt] (-0.2,-1.15) .. controls (0.6,0.3) .. (wiki.240)
  node[midway, right, ann]{write\\(distill)};
% --- read path (retrieval to agent) ---
\node[fold, fill=gray!15] (agent) at (0,3.55) {\textbf{EOP-phase agent}};
\draw[->, black!60, line width=0.6pt] (wiki.north)--(agent.south)
  node[midway, right, ann]{read\\(hybrid $+$ traverse)};
\end{tikzpicture}%
}
\caption{\textbf{The centralized--distributed knowledge layer.} \emph{Distributed local memory}
stores each learning as a note in the code folder it describes (a naturally
tree-structured, human- and agent-navigable store); a \emph{global wiki} abstracts
transferable knowledge and cites those notes, with bidirectional
\emph{document-level} links (dashed). An \emph{entity graph} over models, features,
metrics, and defects adds \emph{entity-level} links (dotted) spanning both tiers. A
proactive \emph{write} path distills modeling and execution learnings; a selective
\emph{read} path retrieves the few relevant to the current EOP phase by hybrid
search plus graph/tree traversal.}
\label{fig:knowledge}
\end{figure}
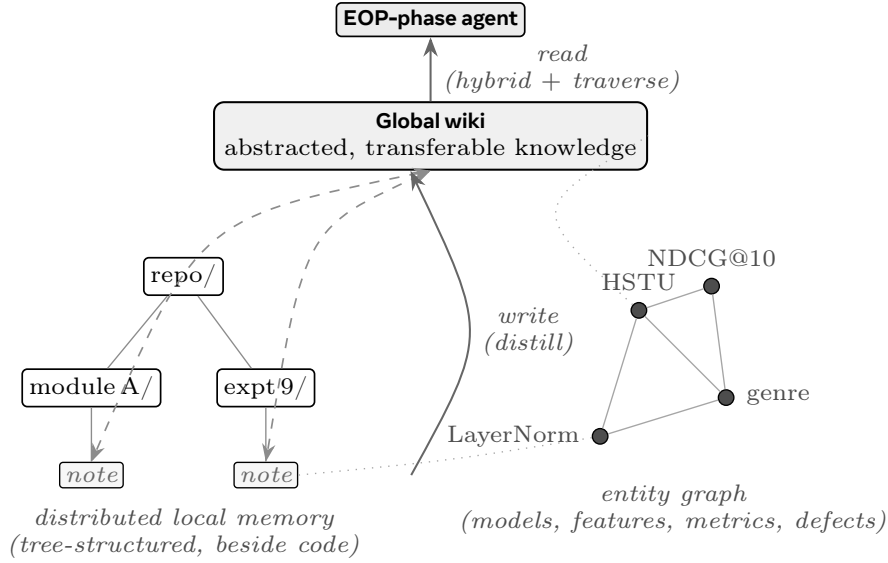

The knowledge layer (Section~\ref{sec:knowledge}) places \re{} in the recent
\emph{agentic-memory} line---agents that autonomously write, organize, and retrieve
their own experience as atomic, interlinked notes or reusable workflows
\citep{xu2025amem, wang2024awm, chhikara2025mem0}---and each ingredient has
precedent in isolation: distilling lessons from successes \emph{and} failures
dates to ExpeL \citep{zhao2024expel}; Agent~KB curates a central store of
workflow- and execution-level patterns with phase-selective retrieval
\citep{tang2025agentkb}; Dolphin retains previously ineffective ideas to veto
near-duplicate proposals \citep{yuan2025dolphin}; industrial recommender agents
keep a persistent journal or insights database of past experiments
\citep{wang2026selfevolvingrec,kumar2026rea}; write-time contradiction
handling appears as memory-update operations and temporal edge invalidation
\citep{chhikara2025mem0, rasmussen2025zep}; and shipped coding products index
human-authored instructions by the repository tree under a most-local-wins
precedence \citep{anthropic2026claudecode}. Unlike \emph{hierarchical}
memory that stacks several levels of semantic abstraction inside a single
index-linked store \citep{sun2026hmem}, our two tiers are distinguished by
\emph{locus}---a centralized wiki versus distributed, code-resident notes---and by
exposing \emph{two} orthogonal access paths (document- and entity-level), not merely
by depth of abstraction.  The \emph{combination} is, to our knowledge, without
prior instance in research systems or products: (i) agent-authored learnings
committed \emph{into the experiment and module folders they describe}, the
repository tree serving as the index; (ii) a declared \emph{central} single
source of truth whose precedence is the inverse of the most-local-wins
convention; (iii) an entity graph spanning both tiers, the substrate on which
contradictory claims meet; and (iv) a standing reconciliation agent that
adjudicates conflicts into condition-scoped conclusions, demotes but back-links
superseded notes, and escalates undecidable conflicts to human review.  We
present this implemented combination as a system-design contribution.  Its
use in the reported deployment is part of the end-to-end evaluation; without
a memory-on/off ablation, however, the observed model gains cannot be
attributed to the knowledge layer alone.  Our instantiation
(Figure~\ref{fig:knowledge}) is two-tier and code-resident, with a \emph{structure},
a \emph{write} path, and a \emph{read} path.

\textbf{Structure (two tiers $+$ a graph).} \emph{Local memory} is distributed:
every learning is a small, self-contained note committed \emph{into the module or
experiment folder it describes}, so the archive mirrors the repository tree and is
navigable by the same \texttt{ls}/\texttt{grep}/path operations humans and agents
already use---no separate database to query, and the note sits exactly where a
reader working on that code will look. A \emph{global wiki} sits above the tree and
distills the cross-cutting, transferable knowledge, treating the local notes as its
worked examples: each wiki entry \emph{cites} the notes it generalizes and each note
\emph{back-links} to the entries that subsume it (bidirectional
\emph{document-level} links). Over both tiers we maintain an \emph{entity graph}
whose nodes are recurring artifacts---models, architectural components, datasets,
metrics, and named defects---and whose edges connect every place an entity is
mentioned; this yields \emph{entity-level} links that let a query on, say,
\texttt{LayerNorm} or \texttt{genre} reach every related note, wiki entry, and
sibling entity at once. Document-level and entity-level links together give agents
two orthogonal, cheap discovery paths over the same content. A learning is written
at the leaf where it arose and abstracted into the wiki as it generalizes, but the
leaf note is kept as that entry's cited evidence, so an agent meets a lesson where
it works rather than recalling it from one ever-growing global memory. Concretely,
wiki entries record what generally helps or hurts and the reusable fix-instructions
and runbooks distilled from the notes they cite.

\textbf{Write (proactive extraction).} Rather than wait to be queried, the
framework proactively spawns a subagent to distill learnings from three sources. From
each experiment's \emph{results and analysis} it records \emph{modeling}
learnings---what helped, what hurt, and why (e.g.\ that an explicit time-decay
add-on is redundant given HSTU's built-in time bias). From the runtime's
review--repair (Dual) step (Section~\ref{sec:metameta}) it records \emph{execution}
learnings: when cross-checking catches a recurring defect---a missing layer norm, a
silently broken gradient path---it is promoted to an explicit, reusable instruction
(``insert the normalization layer after \dots''). From the run's own operational
incidents---failed or stalled jobs and the fixes that unblocked them---it records the
operational side of execution: training and evaluation infrastructure,
environment and configuration fixes, and historical issues, kept as runbooks so that
a setup problem solved once is not re-diagnosed in a later iteration. Each learning
is written as a local note in the relevant folder; the subagent then updates the
affected wiki entries and refreshes the entity graph, rather than appending to one
monolithic log.

\textbf{Read (recall and selection).} At each EOP phase the framework gathers
candidate learnings through three complementary channels---\emph{hybrid}
keyword-and-embedding retrieval, entity-graph traversal from artifacts named in the
current context, and direct folder-tree lookup for the modules in scope---and a
subagent then selects the subset genuinely relevant to the current phase, injecting
only those notes, wiki entries, and instructions. This preserves the per-step
context discipline of the EOP (Section~\ref{sec:eop})---the live phase sees the few
lessons that matter, not the whole archive---while the wiki supplies the general
principle and its linked notes supply the concrete precedent, surfacing the right
hard-won finding at the right moment.

\textbf{Consistency (conflict resolution).} Because learnings accrue over a long
run, two notes can disagree---a fix that helped early may be contradicted once a
later experiment isolates its true cause. We resolve this by treating the \emph{centralized}
global wiki as the layer's \textbf{single source of truth}: local notes are subordinate
evidence, and the standing invariant is that no local note may contradict the wiki. A
dedicated reconciliation agent runs \emph{actively}---not only at write time---to discover
and settle such conflicts. Discovery is cheap because contradictory notes attach to the same
entity-graph nodes, so competing claims about, say, \texttt{time-decay} or \texttt{LayerNorm}
are surfaced rather than left to diverge unseen; every note is append-only and stamped with
its provenance (the iteration, EOP phase, and metric delta that produced it), which gives the
agent the evidence to adjudicate. Reconciliation folds the disagreement back into the
authoritative wiki entry---where possible as a \emph{scoped} conclusion (the condition under which each
finding holds) rather than a single flat verdict---giving precedence to the more recent,
higher-evidence note while the superseded note is demoted but back-linked, so a later agent
can still see \emph{why} a plausible move was abandoned. A conflict the evidence \emph{cannot}
settle---two well-supported findings that genuinely contradict---is not forced into a verdict
but \textbf{flagged for human review}, keeping the autonomous loop self-consistent exactly where its
evidence runs out. The global wiki is thus not merely an index but the authoritative record
against which all distributed local memory is to be reconciled.

\section{Detailed HSTU Case-Study Findings}
\label{sec:appendix-hstu}
This appendix collects the agent's two diagnostic studies, its
evolution-trajectory details, its largest-win side-feature (\textsc{UDK})
analysis, and its full catalogue of negative results, all referenced
from the case study (Section~\ref{sec:results}).

\subsection{Diagnostic-driven research: failure overlap}
\label{sec:diagnostics}

The most transferable methodological artifact of the study is a diagnostic the
agent computed in iter~6 and revisited in iter~12.

\paragraph{The diagnostic.}
For a user with sequence $[i_1,\dots,i_n]$, the standard leave-one-out task
predicts the held-out $i_n$ from the prefix $[i_1,\dots,i_{n-1}]$. Define the
\emph{failure set} $F_T=\{u:\mathrm{rank}_u(i_n)>10\}$. For the \emph{same}
users, re-evaluate the \emph{frozen} model at an internal position---predict
$i_{n-1}$ from $[i_1,\dots,i_{n-2}]$---to get $F_S$, and compute the overlap
\[
O(F_T,F_S)=\frac{|F_T\cap F_S|}{|F_T|},
\]
the fraction of last-position failures that also fail one position earlier.
A high $O$ means failures \emph{persist} across adjacent positions and are
therefore properties of the user/history rather than of the specific held-out
item---i.e.\ addressable by training-side changes.

\paragraph{Why it is leak-free.}
The diagnostic exposes no test label: (1)~each user's test item $i_n$ is
excluded from training; (2)~$i_{n-1}$ is an ordinary training target, so $F_S$
is measured on a seen label---which is why it is used only to locate persistent
failures and is never reported as a metric; and (3)~computing $F_S$ only re-runs
the frozen model with a different \texttt{ignore\_last\_n} argument. It is a
safe post-hoc analysis computable against any checkpoint. The seen label also
makes the diagnostic conservative: it helps the model at $i_{n-1}$---the same
effect that inflates naive multi-position evaluation
(Appendix~\ref{sec:negatives})---so it biases $F_S$ toward fewer failures and
$O$ downward. A high $O$, users who fail even on a label seen in training, is
therefore strong evidence of persistent difficulty, whereas a low $O$ would be
inconclusive.

\paragraph{What it showed.}
On a 0.2127 checkpoint of the 0.2140 LARGE stack, \textbf{75.2\%} of last-position failures
(ML-20M) overlapped 2nd-last failures and \textbf{88.8\%} overlapped the union of
the 2nd/3rd/4th-last; per-item failure rates correlated $r\!\approx\!0.60$ across
positions. A complementary stratification on ML-32M showed failures dominated by
\emph{tail items} (tail HR@10 $3.1\%$ vs.\ head $35.6\%$, an $11.5\times$ gap)
and, counter-intuitively, by \emph{long-history} users ($h\!\geq\!100$: HR@10
$22.2\%$ vs.\ $h\!<\!20$: $44.5\%$). The agent read this as $\sim$+1--2\%
recoverable headroom from training-side changes and used it to motivate both the
(failed) reweighting losses and the (successful) side-feature direction.

\paragraph{The proposed intervention: training-side boost-last-$K$.}
The diagnostic's natural intervention is to \emph{upweight} supervision at the
internal positions where failures persist, without ever touching the held-out
target. With $w_t$ the existing per-position weight (already including the
time-decay multiplier),
\[
w_t' = w_t\cdot\bigl[1+(\beta-1)\,\mathbf{1}\{\,n-K-1\le t< n-1\,\}\bigr],
\]
with $K\in\{1,3,5\}$ and $\beta\in\{1.5,2.0,3.0\}$. This supervision-reweighting
variant is distinct from the inference-time last-$K$ boosting, which gave no
gain (Table~\ref{tab:iters}, iter~12). It is a one-function,
two-parameter change that the agent implemented and smoke-tested without human
help; its runs did not finish within the study, so no result is reported. The point for auto-research agents is
the \emph{closed loop}: an analytical observation the agent produced itself led,
without explicit prompting, to a principled, code-clean intervention.

\subsection{Evolution-trajectory details}
\label{sec:appendix-traj}

\paragraph{Stage~I --- compose a strong model (\textsc{SYNAPSE}).}
The agent's most architectural work composes \textsc{SYNAPSE} on the HSTU backbone
from three known-good blocks: \textbf{SSD-style tiered compression} (logged as
input compression, IC), an $O(N)$ state compression of long history in the spirit
of structured state-space duality \citep{dao2024mamba2}; \textbf{PRISM}
user-conditioning, a feature-wise FiLM modulation $\gamma\odot\text{item}+\beta$
\citep{perez2021film}; and a \textbf{multi-token scoring head} that pools the
sequence into a recent-intent and a long-term-preference token, scored by summed
inner products that---under per-facet normalization---collapse to a single ANN
call (pooling-by-attention \citep{lee2019settransformer}, in the multi-interest
\citep{li2019mind} and multi-vector \citep{khattab2020colbert} spirit). None is
new; their tuned \emph{composition} is the agent's strongest modeling result,
lifting \ndcg{} from 0.2098 to \textbf{0.2161} (+3.0\%; the SSD$+$PRISM stack alone
gives 0.2140). Stage~I also groups the agent's boldest bet---RLHF-style
preference optimization---and all variants \emph{hurt}, DPO collapsing \ndcg{} by
$35.9\%$ relative to its own reference as preference accuracy approached 1.0
(consistent with reward over-optimization; Appendix~\ref{sec:negatives}).

\paragraph{Stages~II--III --- refine, then push the ceiling.}
After Stage~I, returns sharply diminish (Table~\ref{tab:iters}): a
FLUID time-decay add-on---an exponential decay on the real elapsed gap between
interactions---only ties \textsc{SYNAPSE} (HSTU already carries a learnable
relative time-and-position bias), while focal/tail and multi-position auxiliaries
land within noise; the one clean sweep win is a \emph{BASE} dropout reduction from
0.2 to 0.1, giving \ndcg{}~\textbf{0.1948} ($+2.80\%$). Stage~III then pushes the
ceiling by any means: a leak-free \textsc{UDK} genre side-feature, fine-tuned from
the 0.2140 stack rather than \textsc{SYNAPSE}, is the largest
\emph{numeric} result (\ndcg{}~\textbf{0.2192}, $+4.48\%$) but \emph{saturates}---genre,
year, and popularity all plateau at ${\approx}0.219$, and position-averaged test-time
augmentation and inference-time last-$K$ boosting give no gain---exposing an
apparent $0.22$ ceiling
(Appendix~\ref{sec:win}). Genuinely novel methods (scaling, time-aware
mixup, a penultimate-position auxiliary, a test-time self-correction reranker;
iter~12) mark the study's frontier; their runs did not finish within the study.

\paragraph{The inversion.}
The arc inverts the intuition that bolder ideas pay more: the only architectural
work (Stage~I) gave a solid but bounded $+3.0\%$, the boldest objective bet collapsed
$-35.9\%$, and the largest \emph{number} came from the cheapest, least novel
move---a metadata feature that then saturated, leaving real novelty unproven. Our
reading: a contemporary auto-research agent is most reliable at disciplined
execution, failure diagnosis, and pragmatic information-adding, and least reliable at
the modeling novelty we most want from it.

\subsection{The largest win: leak-free side features}
\label{sec:win}

We frame this side-feature as the ML-20M instantiation of \textsc{UDK}
(Universal Distance Kernel), the generic categorical side-feature kernel the agent
proposed: on ML-20M its channels are genre, year, and popularity (here in a simple
additive form; the kernel's gated multi-channel form generalizes to richer
metadata), and the genre channel alone yields the largest \emph{numeric} win.
Adding metadata is one of the oldest levers in recommendation, so we treat the
genre side-feature less as a modeling contribution than as a test of whether the
agent can wield an \emph{expected} lever correctly---and the interesting content is
the leak it had to defeat and the ceiling it exposes. It is at once the biggest
\emph{numeric} improvement and a cautionary tale about leakage. A learnable
embedding table $E_g$ maps an item's genres to a vector, fused additively
\emph{on the encoder input only}:
\[
\tilde{x}_t = x_t + \alpha\cdot\mathrm{meanpool}\!\big(E_g[\,\mathrm{genres}(i_t)\,]\big),
\]
with the supervision computed from the \emph{raw} item embeddings so positives and
sampled negatives are treated symmetrically. The first implementation (v1) instead
injected genre into the item-embedding lookup used by the loss, so positives
carried genre information that negatives did not; training loss collapsed to zero
and eval appeared to spike (0.2163) before degrading---a textbook feature leak the
agent detected and corrected. The leak-free v2 recovered the win
(Table~\ref{tab:genre}): robust to $\alpha\in[0.05,0.20]$, with from-scratch
training lagging far behind fine-tuning. The subsequent saturation across
genre/year/popularity (iter~11) suggests the side-feature acts less like new
information and more like a \emph{structured regularizer} that nudges the encoder
toward semantically coherent neighborhoods---consistent with the iter-8 finding
that the model ``gets the genre right but the item wrong.''

\begin{table}[t]
\centering
\small
\begin{tabular}{lccc}
\toprule
\textbf{Variant} & $\alpha$ & \textbf{\ndcg{}} & \textbf{$\Delta_p$} \\
\midrule
Paper baseline            & ---  & 0.2098 & --- \\
PRISM+IC stack (ref)      & ---  & 0.2140 & +2.0\% \\
\midrule
Genre FT (leak-free)      & 0.05 & \textbf{0.2192} & +4.48\% \\
Genre FT (leak-free)      & 0.10 & \textbf{0.2192} & +4.48\% \\
Genre FT (leak-free)      & 0.20 & 0.2191 & +4.43\% \\
Genre from-scratch$^*$    & 0.10 & 0.1911 & --- \\
\bottomrule
\end{tabular}
\caption{Genre side-feature on ML-20M\,$\times$\,LARGE\@. FT = fine-tuning from the
0.2140 stacked checkpoint. $^*$unfinished run, last evaluated at epoch 26; not an endpoint.}
\label{tab:genre}
\end{table}

\subsection{Negative results}
\label{sec:negatives}

We treat negatives as first-class: an auto-research agent that hides them is
withholding most of its information.

\paragraph{The 35.9\% DPO collapse.}
Adapting DPO \citep{rafailov2023dpo} to recommendation---mining the true next item
as ``chosen'' and the base model's top non-target predictions as
``rejected''---collapsed \ndcg{} by \textbf{$-35.9\%$} relative to a verified
0.2191 reference; IPO \citep{azar2024ipo} and SimPO \citep{meng2024simpo} also hurt
(Table~\ref{tab:rlhf}). All three reached preference accuracy $\approx$1.0 while
ranking quality fell---consistent with, though not proof of, reward
over-optimization. The agent's root-cause analysis (iter~8) is the key insight:
the ``rejected'' top-$K$ items share
$\approx$0.236 genre Jaccard with the target, so they are partially co-relevant,
and the preference objective systematically \emph{devalues valid candidates}. A
leak-free re-derivation (rejecting items mined only from internal positions, never
the held-out target) no longer collapsed but produced $\Delta\!\approx\!0$:
preference-pair fine-tuning is structurally mismatched to this retrieval objective.

\begin{table}[t]
\centering
\small
\begin{tabular}{lcc}
\toprule
\textbf{Method} & \textbf{\ndcg{}} & \textbf{$\Delta$ vs.\ ref} \\
\midrule
Reference (no pref.) & 0.2191 & --- \\
DPO   & 0.1405 & $-35.9\%$ \\
IPO   & 0.2012 & $-8.2\%$ \\
SimPO & 0.1798 & $-17.9\%$ \\
\bottomrule
\end{tabular}
\caption{Preference optimization with target-derived negatives (pre-fix
mining) on ML-20M\,$\times$\,LARGE: all methods
hurt downstream \ndcg{} despite near-perfect preference accuracy.}
\label{tab:rlhf}
\end{table}

\paragraph{Focal / tail reweighting.}
Focal loss \citep{lin2017focal} on tail items (iter~7) yielded no gain even after
three bug-fixes: HSTU's sampled-softmax already balances per-item gradients, and an
extra focal multiplier pushes mass away from the dense head---importing a
classification technique into ranking, where labels are not mutually exclusive.

\paragraph{Position-averaged test-time augmentation does not help.}
The agent hypothesized that averaging predictions over recent positions would
help. It does not, and the agent showed why: each user's target \emph{differs} at
each position, so per-user target overlap across positions is 0\%---there is no
shared label to ensemble, and the variance reduction that justifies test-time
augmentation does not apply. Apparent ``gains'' from naive multi-position eval
were leakage (inflating 0.2192 to 0.2517/0.2577 at \texttt{ignore\_last\_n}$=$1/2).
An argument rather than a sweep turned a tempting idea into a negative result;
the one measured run agrees (${\approx}0.2190$ vs.\ 0.2192;
Table~\ref{tab:iters}, iter~10).

\paragraph{Longer is not better.}
Extending sequence length from 500 to 700 (LARGE) lost \ndcg{} at $+35\%$
wall-time, consistent with per-step underfitting and attention dilution.

\paragraph{What the failures share.}
In each case the agent \emph{correctly identified} a technique that works elsewhere
(preference learning, focal classification, ensembling, long context) and
\emph{incorrectly assumed transfer} without re-deriving the assumptions---a known
LLM-proposer failure mode (over-weighting surface similarity). Our mitigation is
procedural: the loop \emph{requires} a post-hoc diagnosis of every negative before
the next proposal, which is exactly what surfaced the leakage explanations above.

\section{Agent Behavior and Extended Discussion}
\label{sec:appendix-behavior}
This appendix expands the behavioral analysis and discussion of the deployment
case study (Section~\ref{sec:setup}).

%% ============================================================
\subsection{Agent Behavior Analysis}
\label{sec:behavior}

Beyond the metrics, the trajectory reveals \emph{how} the agent behaves, which is
what distinguishes research from hyperparameter search.

\paragraph{Self-correction.}
The agent found and fixed bugs in its \emph{own} code before trusting results:
three bugs in the focal-loss implementation (a tail threshold collapsing onto
zero-frequency padding, un-normalized focal weights amplifying the effective
learning rate, and an un-normalized tail-$\alpha$), each fixed and guarded by unit
tests; one label-leakage bug in the genre feature (iter~9); and one
training-overlap flaw in the preference-pair mining (iter~4). Each fix was
verified before re-running. Notably, the genre and DPO bugs surfaced not from a
failed unit test but as \emph{anomalous metric trajectories}---a training loss
collapsing toward zero while validation \ndcg{} fell---that the next reasoner
call could read off the leaderboard and act on. This argues that the persistent
leaderboard and per-step monitoring are as load-bearing as the LLM reasoning:
they turn silent correctness failures into observable signals.

\paragraph{Evidence-based rejection.}
Of the major directions the agent pursued, a substantial fraction were ultimately
\emph{rejected} on evidence---focal loss, all three preference-optimization
variants, position-averaged test-time augmentation, and long-context---rather than quietly dropped.
This rejection behavior, and the explicit root-cause analysis attached to each, is
what we would want from a careful collaborator and is largely absent from systems
demonstrated only on positive results.

\paragraph{Calibration from failure.}
The early reproduction gap (literature-claimed +4--8\% gains failing to
materialize) propagated forward as a $\sim$50\% discount on subsequent
expected-impact estimates, producing more realistic experiment designs in later iterations.
The agent's own effort estimates, by contrast, remained optimistic by
$\approx$2--3$\times$ throughout.

%% ============================================================
\subsection{Discussion}
\label{sec:discussion}

\paragraph{What worked.}
(i)~\emph{Parallelism}: maintaining many concurrent training jobs behind a single
monitor was the largest multiplier on effective speed. (ii)~\emph{Leak-aware
evaluation}: separating full-test from subset eval prevented chasing noise, and the
leak-free diagnostics (Sections~\ref{sec:diagnostics}--\ref{sec:negatives})
repeatedly caught subtle data leaks a metric-only loop would have rewarded.
(iii)~\emph{Copy-on-modify and checkpoint mirroring}: hard isolation of
per-iteration code and redundant checkpoint preservation made the long run robust
to pod evictions.

\paragraph{What still needs humans.}
(i)~The agent's first instinct is almost always to \emph{stack} another add-on
rather than to \emph{replace} the approach; steering toward novelty was necessary.
(ii)~Its effort and lift estimates are optimistic. (iii)~It cannot yet recognize
when a whole direction (preference optimization) is structurally wrong without
first burning an iteration, nor when to request more compute. (iv)~It produced no
genuine architectural novelty within the study; every win was a composition---and,
consistent with that, the selection the system \emph{actually runs} is an LLM critic, not
the MAP-Elites/best-$K$ population archive that its design documents envision but
have not yet implemented.

\paragraph{What we would change.}
The study's one genuinely novel artifact---the failure-overlap$\rightarrow$boost-last-$K$
program---came from \emph{analysis}, not generation. In this study, the agent
readily generated plausible add-ons; closing the loop from an observation to a clean
intervention was the harder, more valuable step, and we would budget the agent's effort
toward diagnosis accordingly. More broadly, the value of an auto-research agent on
a hard problem is concentrated in its diagnostic steps.

\end{document}